\documentclass[journal]{IEEEtran}
\usepackage{multirow}
\usepackage{amsthm}
\usepackage{amsfonts}
\usepackage{color, soul}
\usepackage{hyperref}
\hypersetup{
    colorlinks=true,
    linkcolor=blue,
    filecolor=blue,
    urlcolor=red,
    citecolor=black,
}
\theoremstyle{plain}

\newtheorem{assumption}{Assumption}

\usepackage{cite}

\ifCLASSINFOpdf
    \usepackage[pdftex]{graphicx}
\else
\fi
\usepackage{amsmath}
\usepackage{algorithm}
\usepackage{algorithmic}

\usepackage{array}
\ifCLASSOPTIONcompsoc
  \usepackage[caption=false,font=normalsize,labelfont=sf,textfont=sf]{subfig}
\else
  \usepackage[caption=false,font=footnotesize]{subfig}
\fi

\usepackage{stfloats}
\begin{document}
\title{Learning the Precise Feature for Cluster Assignment}

\author{Yanhai~Gan,
        Xinghui~Dong,
        Huiyu~Zhou,
        Feng~Gao,
        and~Junyu~Dong% <-this % stops a space
\thanks{This work was supported by the National Key Research and Development Program of China under Grant 2018AAA0100602. The work of Xinghui Dong was supported by the Young
Taishan Scholars Program under Grant tsqn201909060. (Corresponding author: Junyu~Dong.)}
\thanks{Yanhai~Gan, Xinghui~Dong, Feng~Gao, and Junyu~Dong are with the School
of Computer Science and Technology, Ocean University of China, Qingdao 266100, China (e-mail: see \url{http://ai-ouc.cn/people.html}).}% <-this % stops a space
\thanks{Huiyu~Zhou is with the School of Informatics, University of Leicester, United Kingdom (e-mail: hz143@leicester.ac.uk).}}% <-this % stops a space


\markboth{Journal of \LaTeX\ Class Files,~Vol.~14, No.~8, August~2015}%
{Shell \MakeLowercase{\textit{et al.}}: Bare Demo of IEEEtran.cls for IEEE Journals}



\maketitle

\begin{abstract}
Clustering is one of the fundamental tasks in computer vision and pattern recognition. Recently, deep clustering methods (algorithms based on deep learning) have attracted wide attention with their impressive performance. Most of these algorithms combine deep unsupervised representation learning and standard clustering together. However, the separation of representation learning and clustering will lead to suboptimal solutions because the two-stage strategy prevents representation learning from adapting to subsequent tasks (e.g., clustering according to specific cues). To overcome this issue, efforts have been made in the dynamic adaption of representation and cluster assignment, whereas current state-of-the-art methods suffer from heuristically constructed objectives with representation and cluster assignment alternatively optimized. To further standardize the clustering problem, we audaciously formulate the objective of clustering as finding a precise feature as the cue for cluster assignment. Based on this, we propose a general-purpose deep clustering framework which radically integrates representation learning and clustering into a single pipeline for the first time. The proposed framework exploits the powerful ability of recently developed generative models for learning intrinsic features, and imposes an entropy minimization on the distribution of the cluster assignment by a dedicated variational algorithm. Experimental results show that the performance of the proposed method is superior, or at least comparable to, the state-of-the-art methods on the handwritten digit recognition, fashion recognition, face recognition and object recognition benchmark datasets.
\end{abstract}


% \fi
\IEEEpeerreviewmaketitle

\section{Introduction}
\label{intro}
\IEEEPARstart{D}{eep} neural networks (DNNs) have demonstrated their powerful ability in computer vision tasks, such as object detection~\cite{redmon2018yolov3}, classification~\cite{chen2017dual}, instance segmentation~\cite{he2017mask} and scene understanding\cite{DBLP:journals/pami/ShelhamerLD17}. However, the training of a robust and efficient DNN generally requires a large amount of annotated data. For example, over one million labeled images divided into 1000 categories are contained in the ImageNet dataset~\cite{deng2009imagenet}, and more than 375 million noisy labels are assigned to 300 million images in the JFT-300M~\cite{DBLP:conf/iccv/SunSSG17}. As we have known, it is very time-consuming and labor-expensive to collect such a large-scale annotation set~\cite{liu2018perception, DBLP:conf/eccv/LinMBHPRDZ14}. On the other hand, large quantities of images, videos, and other types of data are being produced every day. It is indeed impractical to manually annotate these data. Therefore, it is crucial to develop methods that can automatically exploit knowledge from unlabeled data.

Neuroscientists have empirically proven that the naturalistic visual experience plays a fundamental role in developing a powerful visual system \cite{wood2018development,arcaro2017seeing}. This indicates that unsupervised learning happens constantly in the human perceptual system. Normally, unsupervised learning methods model the underlying structure or distribution of the input data without annotation~\cite{DBLP:journals/tcsv/ZhouXCYC19}. As an unsupervised learning paradigm, clustering aims to divide the input data into a set of clusters according to the distributional attributes of the data~\cite{6817617, DBLP:conf/cvpr/ZhaoCM18, DBLP:journals/tcyb/LiC20, DBLP:journals/tcyb/PangXNL20}. However, standard clustering algorithms usually depend on some predefined distance metrics which are usually difficult to identify for high dimensional data~\cite{DBLP:journals/sac/Luxburg07, DBLP:conf/nips/GomesKP10, gowda1978agglomerative, DBLP:journals/pr/0010ZW13}. Furthermore, the time complexity of standard clustering algorithms will dramatically increase when large-scale datasets are encountered~\cite{DBLP:conf/iccv/0004HNCW17}.

To mitigate these issues faced by standard clustering methods, researchers first embedded the input data into a new low-dimensional space and then implemented a standard clustering method in the embedding space \cite{DBLP:journals/corr/KingmaW13, tacchetti2018trading, ren2018cross, DBLP:conf/cvpr/JenniF18}. In this two-stage scheme, representation learning is agnostic to the following clustering task, and thus can hardly produce the representative features for a specific task. Therefore, some efforts have been made to dynamically adapt the representation and cluster assignment~\cite{DBLP:conf/cvpr/ZhouHF18, DBLP:conf/cvpr/LinCCC18, DBLP:conf/nips/ChenA018, xie2016unsupervised, Chang2017Deep, DBLP:journals/access/ZhouZ19b, DBLP:conf/cvpr/ZhangLYQZ0L19}. These methods generally assume that the label of each cluster can be used as supervisory signals to learn representations and in turn the representations will be beneficial to instance clustering. Consequently, the core idea of these methods is to apply a strategy to alternating between representation learning and clustering~\cite{yang2016joint, Chang2017Deep, DBLP:conf/iccv/DizajiHDCH17, shaham2018spectralnet, DBLP:journals/access/ZhouZ19b, DBLP:conf/cvpr/ZhangLYQZ0L19}. Although such methods have produced promising results, the heuristically constructed objectives lack a principled characterization of goodness of deep clustering, thus making the good performance of deep clustering models customized~\cite{9122459, 9208789}.

Rather than conducting representation learning and clustering separately, humans tend to take into account these two tasks as a whole. For instance, one is likely to perform clustering according to the gender when he/she is asked to divide his/her colleagues into two groups. Nevertheless, he/she can also consider other characteristics, such as position, age and income, for clustering in terms of desired groups. That is to say, humans tend to discover the exactly matched features with regard to the desired number of groups and perform clustering accordingly. Inspired by this, we define the objective of deep clustering as finding a precise feature as the cue for cluster assignment. This objective provides a fresh avenue of exploration -- how to optimally select a deep clustering architecture and how to best design the optimization objective. Meanwhile, this objective encourages the development of solutions for dealing with general-purpose clustering tasks.

Further insight into the decision-making mechanism of humans can provide us with the intuition that representation learning and clustering are collaborative tasks and expected to work together to produce the desirable results. In the proposed framework, we integrate representation learning and clustering into a single pipeline for joint optimization instead of alternating between them as in previous methods \cite{yang2016joint, Chang2017Deep, DBLP:conf/iccv/DizajiHDCH17, shaham2018spectralnet, DBLP:journals/access/ZhouZ19b, DBLP:conf/cvpr/ZhangLYQZ0L19}. To the best of our knowledge, this is the first attempt which essentially couples representation learning and clustering. The main contributions of this work are summarized as follows:

\begin{itemize}

  \item A principled deep clustering objective is proposed to find a precise feature as the cue for cluster assignment.

  \item A necessary and sufficient condition is postulated to enable a solution for accomplishing the stated objective.

  \item A general-purpose deep clustering framework that couples representation learning and clustering is introduced.

  \item The state-of-the-art clustering results obtained using the proposed framework on several public datasets provide the other researchers a set of benchmarks.

\end{itemize}

The remaining of this paper is organized as follows. In Section \ref{relatedworks}, we review the existing related work. The core ideas of the proposed framework are introduced in Section \ref{sec:method}. In Section \ref{sec:exp}, we present the experimental results. Finally, our conclusions and future work are given in Section \ref{sec:Con}.

\section{Related Work}
\label{relatedworks}
Standard clustering algorithms (such as K-means~\cite{6817617}, spectral clustering~\cite{DBLP:journals/sac/Luxburg07}, Gaussian
mixture models~\cite{DBLP:journals/pami/X00}) tend to encounter difficulties when dealing with high-dimensional and large-scale datasets~\cite{DBLP:journals/tnn/WangCZHY19, DBLP:journals/tip/TarunB18, DBLP:journals/cacm/Domingos12}. In this regard, many kinds of two-stage methods are explored~\cite{DBLP:conf/aaai/TianGCCL14, xie2016unsupervised, DBLP:conf/ijcai/PengXFYY16, DBLP:journals/kbs/RenWLX20}. These methods first projected the data into a low-dimensional manifold, and then applied standard clustering algorithms~\cite{DBLP:journals/csr/BouwmansSMZBF18, DBLP:conf/iccv/Lowe99, pathak2016context}. However, these methods normally require domain-specific architectural deliberation in order to learn discriminative representations \cite{DBLP:conf/eccv/NorooziF16, noroozi2017representation, arandjelovic2017look}. Although such deliberation is necessary for obtaining the competitive clustering performance, it is harmful to choosing a suitable architecture for a given task. It makes state-of-the-art deep clustering architectures become increasingly domain-specific \cite{tacchetti2018trading, ren2018cross, DBLP:conf/cvpr/JenniF18, DBLP:journals/tmm/WangXKY19}. In addition, after being optimized in the first stage, the learned representation is fixed, so it cannot be further improved to obtain better performance in the clustering stage.

%In order to achieve dynamic adaptation of representation and cluster assignment, some efforts have been made to apply an algorithm similar to Expectation Maximization (EM) to alternately estimate cluster assignment and update the representation
Recently, some efforts have been made in the dynamic adaptation of representation and cluster assignment~\cite{DBLP:conf/cvpr/ZhouHF18, DBLP:conf/cvpr/LinCCC18, DBLP:conf/nips/ChenA018, xie2016unsupervised, Chang2017Deep, DBLP:journals/access/ZhouZ19b, DBLP:conf/cvpr/ZhangLYQZ0L19}. As an early work, deep embedded clustering (DEC)~\cite{xie2016unsupervised} improves the clustering using an unsupervised algorithm that alternates between two steps: 1) computing a soft assignment between the embedded points and the cluster centroids, 2) updating the deep mapping and refine the cluster centroids by learning from current high confidence assignments using an auxiliary target distribution. Analogously, Yang {\it et al.}~\cite{yang2016joint} propose a JULE framework that formulates the successive operations in a clustering algorithm as the steps in a recurrent process. One step updates the cluster labels using the current representation while another step updates the representation parameters based on the current clustering results. Lately, Chang {\it et al.} \cite{Chang2017Deep} propose a deep adaptive clustering (DAC) algorithm that recasts the clustering problem into a binary pairwise-classification problem for judging whether or not a pair of images belong to the same cluster. To further utilize the category information, Wu {\it et al.}~\cite{DBLP:conf/iccv/WuLWQLLZ19} develop a deep comprehensive correlation mining (DCCM) method that is trained by selecting highly-confident information in a progressive way.

Kamran {\it et al.} \cite{DBLP:conf/iccv/DizajiHDCH17} introduce a multinomial logistic regression method on top of a multi-layer convolutional autoencoder for the joint learning of representation and cluster assignment. This method was referred to as deep embedded regularized clustering (DEPICT). Similarly, Zhou {\it et al.}~\cite{DBLP:journals/access/ZhouZ19b} form an adversarial deep embedded clustering by combining adversarial auto-encoder and k-means together, where the representation parameters and clustering results are iteratively fine-tuned in a form of self-training after the network has been pretrained. To overcome the shortcomings of traditional spectral clustering, Shaham {\it et al.}~\cite{shaham2018spectralnet} propose a deep learning based method (SpectralNet) that learns a map to embed input data points into the eigenspace of their associated graph Laplacian matrix and then performs the clustering operation. Based on the same inspiration,  Zhang {\it et al.}~\cite{DBLP:conf/cvpr/ZhangLYQZ0L19} combine convolutional networks, self-expression module and spectral clustering module into a joint optimization framework ($S^2ConvSCN$), in which the current clustering results are used to self-supervise the training of the feature learning and the self-expression module.

Although these methods have devoted huge efforts to the dynamical adaptation of representation and cluster assignment, and have produced promising results, they usually employ an alternative optimization strategy for representation learning and clustering. As a result, these methods usually prefer certain datasets and incorporate many exotic designs for learning discriminative features. In contrast, we would like to define the objective of deep clustering in a principle way. Specifically, we are committed to finding a precise feature as the cue for cluster assignment. To this end, we radically integrate the representation learning and clustering into a single pipeline rather than alternating between the two tasks. As a result, we discard those exotic designs for the representation learning, and finally come up with a general-purpose deep clustering framework that can be generalized to common clustering tasks.

As the implementation involves the generative adversarial networks (GANs)~\cite{goodfellow2014generative}, we make a brief introduction to GANs. Most commonly, it is recognized that GANs were proposed by Ian J. et al. in 2014 \cite{goodfellow2014generative}. Compared to the blurry and low-resolution outcome from other generative models \cite{DBLP:journals/corr/KingmaW13,DBLP:conf/cvpr/DosovitskiySB15}, GANs-based methods \cite{radford2015unsupervised,DBLP:conf/iccv/MaoLXLWS17,DBLP:conf/eccv/ZhuKSE16,isola2017image} generate more realistic results. However, training GANs is well acknowledged to be delicate and unstable~\cite{radford2015unsupervised, salimans2016improved, arjovsky2017towards}. The problem is that the JS distance, which is essentially optimized by GANs, is not a continuous loss function on the model's parameters. WGANs cure this problem by continuously estimating the Earth Mover distance~\cite{arjovsky2017wasserstein}. Nevertheless, WGANs sometimes still generate poor samples or fail to converge due to the use of weight clipping to enforce a Lipschitz constraint on the critic. To rescue WGANs from the pathological performance, Gulrajani {\it et al.} \cite{gulrajani2017improved} propose to penalize the norm of the gradients of the critic with respect to its input (WGAN-GP) as an alternative to clipping weight. WGAN-GP performs much better than standard WGANs and enables stable training of a wide variety of GANs architectures with light hyperparameter tuning.

%Even more, the heuristically constructed objective also leads to delicate architecture, which requires elaborate control and cautious hyperparameter-tuning in the training process

\section{The unified deep clustering framework}\label{sec:method}
In this section, we first formulate the objective of deep clustering as finding a precise feature as the cue for cluster assignment. Furthermore, to motivate a paradigm to fulfill this objective, we religiously make a basic assumption about deep clustering. Secondly, we introduce the discipline for constructing the proposed unified deep clustering framework based on the assumption. Finally, we describe the implementation practice of the proposed deep clustering framework.

\subsection{Objective Formulation}
Unlike supervised learning, where the learning objective can be straightforward defined as the closeness between the ground-truth annotation and the prediction \cite{DBLP:conf/nips/ZhangS18, DBLP:conf/iclr/LaineA17}, how to define the objective of unsupervised learning is still an open problem worth exploring~\cite{vincent2010stacked, zhao1506stacked, pathak2016context, noroozi2017representation, arandjelovic2017look, tacchetti2018trading, ren2018cross, DBLP:conf/cvpr/JenniF18}. Generally, the objective is defined to discover the most discriminative features of the data points~\cite{tacchetti2018trading, ren2018cross, DBLP:conf/cvpr/JenniF18}. However, in the deep clustering context, discriminative features can be task-specific. For example, digit and handwriting recognition may require different cues of the input image samples. In other words, the best features for digit recognition are not necessarily suitable for handwriting recognition. As a matter of common sense, digit type is the quite discriminative feature of the handwritten digit dataset. Nonetheless, these digit images may be sampled from different writers. Indeed, if we would like to cluster these digit images according to their handwriting, none of the existing clustering algorithms perform well because their objectives are all dedicated to digit recognition.

To address this issue, we redefine the objective of deep clustering as finding a precise feature as the cue for cluster assignment. This objective encourages the establishment of frameworks for handling general-purpose clustering tasks. These tasks may include the handwriting clustering problem mentioned above, and all that is required is just the prior knowledge of the number of clusters. Please note that, by referring to the precise feature, we mean an exactly matched feature whose possible values can establish a one-to-one relationship with the designated clusters. Therefore, a precise feature accounts for a maximum predictability of the cluster assignment for a given sample. Mathematically, maximization of predictability is equivalent to the minimization of the entropy of a distribution. As a result, an appropriate approach to achieving the defined objective is to minimize the distribution entropy of the cluster assignment of a given sample.

However, it is well known that, when representation and cluster assignment are jointly optimized, direct entropy minimization is prone to getting stuck in the non-optimal local minima during training~\cite{DBLP:conf/iccv/DizajiHDCH17}. The reason is that practical samples usually contain a large number of variables that create many spurious correlations. To avoid the learning method from falling into these spurious correlations (e.g., division of the value space of a real-valued feature), we set up two constraints: 1) there is no empty cluster, 2) the cue feature for cluster assignment is unique-valued in a cluster. Under these two constraints, the learning method will then be guided to select a precise feature as the cue for cluster assignment. All these are further formalized in the following assumption.

\begin{assumption}\label{ass:1}
Minimization of the expectation of the distribution entropy of the cluster assignment is the necessary and sufficient condition for learning a precise feature as the cue for cluster assignment, under two constraints: 1) there is no empty cluster and 2) the cue feature for cluster assignment is unique-valued in a cluster.
\end{assumption}
More explanation of this assumption can be found in the supplemental material. According to Assumption 1, we can endow a deep clustering algorithm with a similar ability to humans by exploiting the most suitable features for different clustering tasks. In the next subsection, we will introduce the framework substantiating this assumption.
\subsection{Framework Design}
\label{sec:1}
\begin{figure*}
\centering
\begin{minipage}{0.34\linewidth}
  \centerline{\includegraphics[width=2.5in]{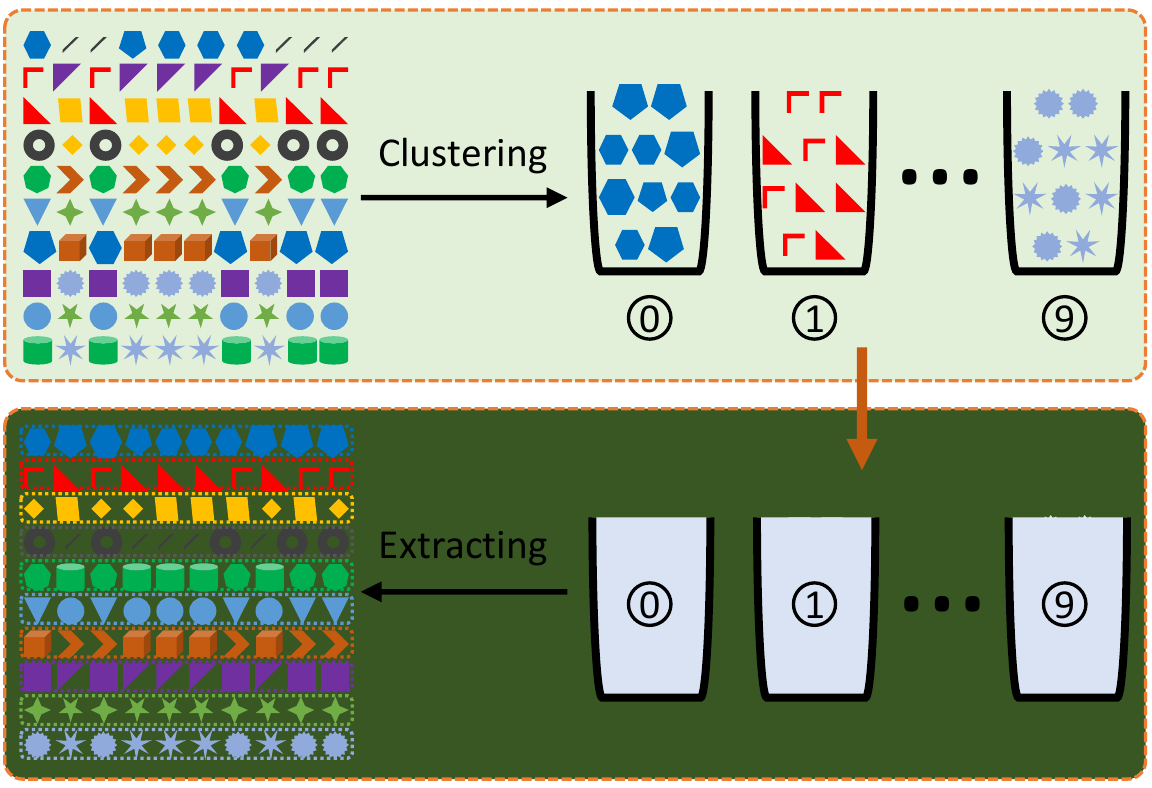}}
  \centerline{(a)}
\end{minipage}
\hfill
\begin{minipage}{0.62\linewidth}
  \centerline{\includegraphics[width=4.56in]{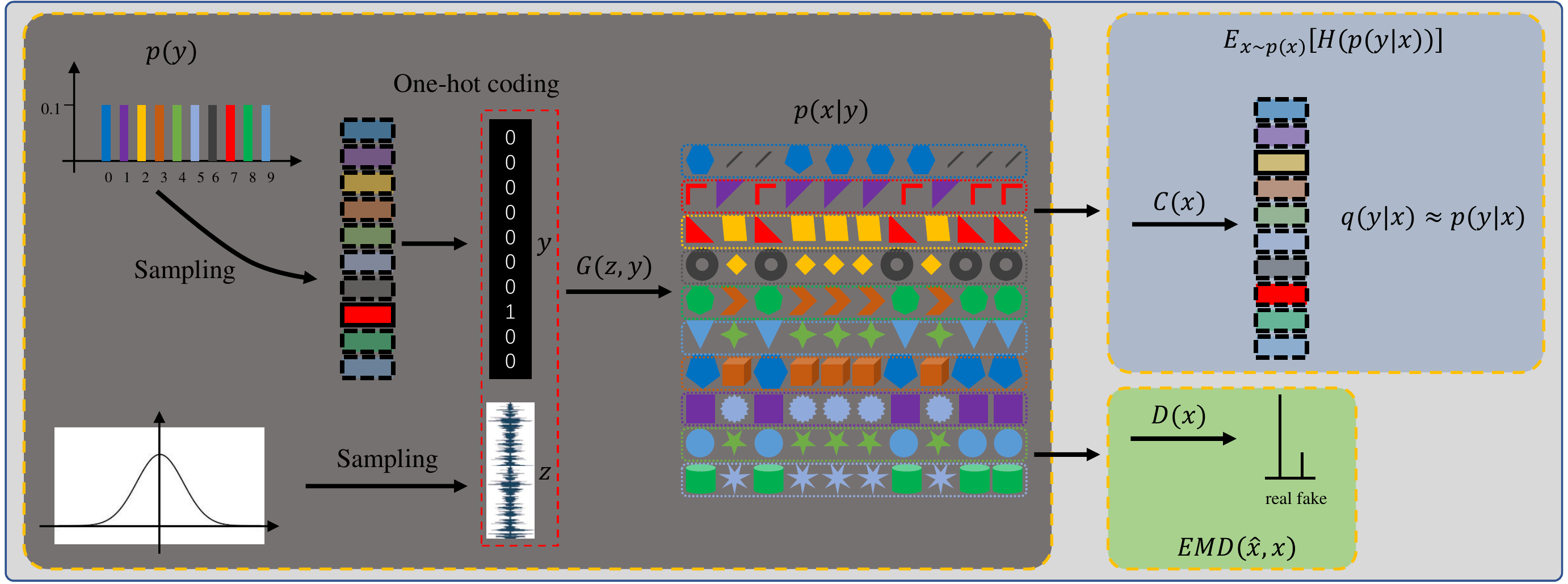}}
  \centerline{(b)}
\end{minipage}
\caption{Illustration of the overall framework. (a) is a simple sketch of the clustering process. In the first row, 100 blocks of 10 colors, 20 shapes and 5 sizes are clustered into 10 baskets. Equivalently, as depicted in the second row, the basket assignment can be derived by extracting blocks from these baskets. In the second row, the baskets are covered in gray because the assignment is unknown until all the blocks are extracted from these baskets. (b) is the flowchart of the proposed framework, where $x$ denotes a sample and $y$ denotes a cluster id. Besides, $z$ is a random variable that obeys a multivariate normal distribution (with covariance matrix being an identity matrix), representing the features independent of $y$. In the framework, $C$ is optimized to estimate the expectation of the distribution entropy of the cluster assignment, and $D$ aims to estimate the Earth Mover distance ($EMD$) between the generated samples and the real samples. Afterwards, $G$ is optimized to minimize the expectation of the distribution entropy of the cluster assignment and $EMD$ simultaneously.}
\label{fig:framework}
\end{figure*}

%\begin{figure}
%  \centering
%  % Requires \usepackage{graphicx}
%  \includegraphics[width=3.5in]{generate.pdf}\\
%  \caption{A simple sketch of the clustering process. In the first row, 100 blocks of 10 colors, 20 shapes and 5 sizes are clustered into 10 baskets. Equivalently, as depicted in the second row, the basket assignment can be derived by extracting blocks from these baskets. In the second row, the baskets are covered in gray because the assignment is unknown until all the blocks are extracted from these baskets}\label{fig:generate}
%\end{figure}

%\begin{figure*}
%  \centering
%  % Requires \usepackage{graphicx}
%  \includegraphics[width=7.1in]{architecture.pdf}\\
%  \caption{The flowchart of the proposed framework, where $x$ denotes a sample and $y$ denotes a cluster id. Besides, $z$ is a random variable that obeys a multivariate normal distribution, which represents the features independent of $y$. In the framework, $C$ is optimized to estimate the expectation of the distribution entropy of the cluster assignment, and $D$ aims to estimate the Earth Mover distance ($EMD$) between the generated samples and the real samples. Afterwards, $G$ is optimized to minimize the expectation of the distribution entropy and $EMD$ simultaneously}\label{fig:corresponding}
%\end{figure*}

First of all, the main inspiration and the overall framework have been illustrated in Fig. \ref{fig:framework}. In Fig. \ref{fig:framework}(a), 100 blocks of 10 colors, 20 shapes and 5 sizes are assigned to 10 baskets. As shown in the bottom of Fig. \ref{fig:framework}(a), the assignment can be derived by fetching blocks from these baskets. If $x$ and $y$ denote a sample and a cluster respectively, the clustering process can be described as assigning $x$ to $y$. Equivalently, this process can be resolved by extracting $x$ from $y$. The inverse formula naturally conforms to a generative paradigm from $y$ to $x$, which inspires us to design the deep clustering framework as shown in Fig. \ref{fig:framework}(b). In Fig. \ref{fig:framework}(b), $G$ is an implicit generative model acting as the process of fetching samples from given cluster ids, $D$ is a discriminator used to estimate the consistency between the generated samples and the real sample, $C$ is a classifier approximating the real posterior distribution of the cluster assignment. In the following, we detail the motivation for each design choice of the framework.

Using $G$ to reversely simulate the clustering process helps to achieve the constraints of Assumption \ref{ass:1}. This is well exemplified in Fig. \ref{fig:framework}(a). In Fig. \ref{fig:framework}(a), if shape is chosen as the cue for basket assignment, there will be at least one basket containing blocks of at least two different shapes (Pigeonhole principle). Subsequently, as shown in the bottom of Fig. \ref{fig:framework}(a), if we use the identity of this basket as the condition for generation, the produced samples will be of only one shape. This is due to the fact that there is no additional information for indicating what shapes exist in that basket. Conclusively, if the cue used for cluster assignment is not an intrinsic feature, but the value space division of a certain feature, the generated samples will drop into a subspace of the original sample space. In Fig. \ref{fig:framework}(b), $D$ is used to ensure the consistency between the generated samples and the real samples. Under the consistency requirement, the learning method has to put samples of the same feature value into one cluster. As a consequence, the second constraint of Assumption \ref{ass:1} can be satisfied.

In this paper, we refer to the features that are independent on each other and essential to composing the sample space as intrinsic features. In this sense, we can say that the generative formulation makes the deep model learn an intrinsic feature as the cue for cluster assignment. For the first constraint of Assumption \ref{ass:1}, we impose it by adopting a uniform prior --  denoted by $p(y)$ in Fig. \ref{fig:framework}(b) -- on the marginal distribution of the cluster assignment. In this way, we virtually assume that samples are evenly distributed across clusters. In practice, the number of clusters is usually predetermined, but the marginal distribution is often unknown. Therefore, the uniform prior is indeed an over implementation of the first constraint, which limits the applicability of the clustering framework. Its specific impact will be further analyzed in the experiment section.

Since the two constraints of Assumption \ref{ass:1} have been satisfied, we are to deduce the entropy minimization objective required by the assumption. Although it is straightforward to perform an entropy minimization in a discriminative model, this is not the case for a generative model, because the distribution of the cluster assignment therein is posterior and always intractable. For this reason, we introduce a variational algorithm for indirect optimization of the distribution entropy of the cluster assignment. Specifically, we first calculate an approximation (output of $C$) of the real posterior distribution, and then induce an upper bound of the expectation of the distribution entropy. Afterwards, the expectation of the distribution entropy of the cluster assignment can be consistently minimized as we continue to lower the upper bound.

As illustrated in Fig. \ref{fig:framework}(b), the conditional distribution implied by $G$ is denoted as $p(x|y)$, the posterior distribution of the cluster assignment is denoted as $p(y|x)$, and the approximation of $p(y|x)$ is denoted as $q(y|x)$. Afterward, the expectation of the cross-entropy between the real posterior distribution and the approximation can be calculated as follows:

\begin{eqnarray}\label{eq:expectation}
  && E_{x\sim p(x)}[H(p(y|x),q(y|x))] \nonumber \\
  &=& -\int p(x)\int p(y|x)\log q(y|x)dydx \nonumber \\
  &=& -\int\int p(x)p(y|x)\log q(y|x)dydx \nonumber \\
  &=& -\int\int p(x,y)\log q(y|x)dydx \nonumber \\
  &=& E_{x,y\sim p(x,y)}[-\log q(y|x)].
\end{eqnarray}
Eq.(\ref{eq:expectation}) pronounces that the expectation of the cross-entropy is equal to the expectation of the negative log-likelihood of the approximation on the joint distribution of $x$ and $y$.

In addition, the expectation of the cross-entropy can be expressed as an addition of two terms:
\begin{eqnarray}\label{eq:convert}
  && E_{x\sim p(x)}[H(p(y|x),q(y|x))] \nonumber \\
  &=& E_{x\sim p(x)}[-\int p(y|x)\log q(y|x)dy] \nonumber \\
  &=& E_{x\sim p(x)}[-\int p(y|x)\log (\frac{q(y|x)}{p(y|x)}p(y|x))dy] \nonumber \\
  &=& E_{x\sim p(x)}[-\int p(y|x)(\log \frac{q(y|x)}{p(y|x)}+\log p(y|x))dy] \nonumber \\
  &=& E_{x\sim p(x)}[\int p(y|x)\log \frac{p(y|x)}{q(y|x)}dy-\int p(y|x)\log p(y|x)dy] \nonumber \\
  &=& E_{x\sim p(x)}[KL(p(y|x),q(y|x))]+E_{x\sim p(x)}[H(p(y|x))] ,
\end{eqnarray}
where $KL(p(y|x),q(y|x))$ represents the Kullback-Leibler divergence between $p(y|x)$ and $q(y|x)$, and $H(p(y|x))$ denotes the distribution entropy of $p(y|x)$. Since $KL(p(y|x),q(y|x))$ is definitely positive, we have the following inequality:
\begin{equation}\label{eq:ineq}
  E_{x\sim p(x)}[H(p(y|x))]\leq E_{x\sim p(x)}[H(p(y|x),q(y|x))] .
\end{equation}
Eq.(\ref{eq:ineq}) indicates that the expectation of the cross-entropy is an upper bound of the expectation of the distribution entropy of the cluster assignment, and the upper bound becomes tight if and only if $KL(p(y|x),q(y|x))$ gets close to zero, which means that $q(y|x)$ is approaching $p(y|x)$ almost everywhere. Therefore, we will consistently minimize the expectation of the distribution entropy of the cluster assignment if we keep the approximation $q(y|x)$ accurate and continue to reduce the cross-entropy $E_{x\sim p(x)}[H(p(y|x),q(y|x))]$.

However, the solution illustrated by Eq.(\ref{eq:expectation}) for the expectation of the cross-entropy involves an expectation on the joint distribution $p(x,y)$ that is implicit. Since direct calculation is not practical, we solve for the expectation by utilizing a Monte-Carlo algorithm. As we encode cluster ids in the one-hot fashion, according to the strong law of large numbers, the following equation can be obtained:
\begin{eqnarray}\label{eq:mc}
  &&E_{x,y\sim p(x,y)}[-\log q(y|x)] \nonumber \\
  &=&\lim_{n\to\infty}-\frac{1}{n}\sum_{i=1}^{n}\sum_{j=1}^{k}y_{ij}\log q(y_j|x_i) ,
\end{eqnarray}
where $y_{ij}$ represents the $jth$ entry of the one-hot coding vector of the cluster id generating $x_i$, $k$ is the number of clusters, and $n$ denotes the number of samples. Eq.(\ref{eq:mc}) enables a Monte-Carlo solution for the expectation by first sampling $y$ from the prior distribution and then sampling $x$ from the likelihood $p(x|y)$ that is implicitly modeled by the generator $G$. The two-stage sampling process is equivalent to sampling $(x, y)$ from their joint distribution $p(x, y)$. In practice, the Monte-Carlo approximation becomes more and more accurate with the value of $n$ getting larger. Here we suppose that $n$ is large enough, and the solution is basically accurate.

In the training stage, $D$ and $C$ are optimized first before each optimization of $G$. It is known that the discriminator $D$ is dedicated to estimating a distance between the generated samples and the real samples~\cite{goodfellow2014generative, zhao2016energy, arjovsky2017wasserstein, gulrajani2017improved}. Because $C$ is used to approximate the real posterior distribution of the cluster assignment of a given sample, the Kullback-Leibler divergence between $p(y|x)$ and $q(y|x)$ will approach zero after $C$ is optimized. This means that $q(y|x)$ asymptotically equals to $p(y|x)$ almost everywhere and the upper bound implied by Eq.(\ref{eq:ineq}) hence becomes tight. In turn, the generative model $G$ is optimized for two tasks: 1) minimization of the distance between the generated samples and the real samples, 2) reducing the upper bound of the expectation of the distribution entropy of the cluster assignment. Consequently, the generative model $G$ learns a mapping between the cluster ids and the samples, where the minimized distribution entropy of the cluster assignment ensures a one-to-one correspondence between the clusters and the discrete values of the cue feature.

In this subsection, we assume that samples and labels are continuous and perform derivation by calculus, whereas it should be noted that the conclusions still hold for discrete variables. In that case, the integration becomes summation and the probability densities become discrete probability masses.

\subsection{Implementation}

First, the uniform distribution $y\sim\mathcal{U}^{int}[1, k]$, which denotes a discrete distribution with the probability mass uniformly distributed on integers in the closed interval $[1, k]$, is employed as the marginal distribution of the cluster assignment to satisfy the first constraint of Assumption \ref{ass:1}. Second, as WGAN-GP~\cite{gulrajani2017improved} achieves much better performance than other generative models \cite{radford2015unsupervised, salimans2016improved, arjovsky2017towards, arjovsky2017wasserstein}, we employ WGAN-GP~\cite{gulrajani2017improved} as the backbone to realize the consistency between the generated samples and the real samples. The consistency condition satisfies the second constraint of Assumption \ref{ass:1}. In the following, we formally present the loss functions for each component (discriminator, estimator, generator) of the framework.

First of all, the loss function of the discriminator $D$ is the same defined as that in WGAN-GP:
\begin{eqnarray}\label{d_loss}
  L_D = E_{x\sim p_g(x)}[D(x)]-E_{x\sim p_r(x)}[D(x)] \nonumber \\
  +\lambda E_{\hat{x}\sim p_{\hat{x}}(\hat{x})}[(\|\nabla_{\hat{x}}D(\hat{x})\|_2-1)^2],
\end{eqnarray}
where $p_{\hat{x}}()$ represents the uniform sampling function which works along the straight lines between a pair of points sampled from both the data distribution $p_r$ and the generator distribution $p_g$, $D(x)$ denotes the output of the discriminator when $x$ is given, and $\lambda$ is a hyperparameter for the gradient penalty term. After the discriminator $D$ has been sufficiently optimized, $L_D$ will be approximately equal to the Earth Mover distance~\cite{gulrajani2017improved} between the generated samples and the real samples.

Second, because the cluster assignment of a given sample obeys a categorical distribution, the estimator $C$ for the posterior distribution of the cluster assignment adopts the conventional cross entropy as the loss function:
\begin{equation}\label{c_loss}
  L_C = -\frac{1}{n}\sum_{i=1}^{n}\sum_{j=1}^{k}y_{ij}\log q(y_j|x_i),
\end{equation}
where $y_{ij}$ denotes the $jth$ entry of the one-hot coding of the cluster id used as the input in $G$ to generate the $ith$ sample, and $q(y_j|x_i)$ stands for the $jth$ output of the classifier when the $ith$ sample is fed as input. The cluster ids fed into the generator are viewed as ground-truth when the cross-entropy loss is calculated, and we refer them as fake labels in the following. It should be noted that the fake labels are autonomously generated rather than being annotated by humans.

Finally, the loss function of the generator $G$ is defined as an addition of two terms:
\begin{equation}
  L_G = L_{reality} + \eta L_{entropy},
\end{equation}
where $L_{reality}$ is the reality term, and $L_{entropy}$ is the entropy minimization term. According to the conventional practice of GANs~\cite{gulrajani2017improved}, we make $L_{reality} = -E_{x\sim p_g(x)}[D(x)]$. Since $L_C$ is defined as the cross entropy by Eq.(\ref{c_loss}), we can readily set $L_{entropy}=L_C$ by referring to Eq.(\ref{eq:mc}). In addition, $\eta>0$ is a trade-off parameter between the reality term and the entropy minimization term. During training, $\eta$ is exponentially increased in a staircase function:
\begin{equation}\label{eq:eta}
  \eta\leftarrow\eta\gamma^{\lfloor t/\tau\rfloor} ,
\end{equation}
where $t$ represents the current training step, and $\tau$ denotes the number of steps for every increase. In this manner, the generator will tend to focus on the entropy minimization objective in the later training stage, in which the quality of the generated samples has been significantly improved.

The training dynamics of these three components are further formalized in Algorithm \ref{algo}, which also declares the configuration of the hyperparameters used in the experiment. After the optimization is completed, the estimator for the posterior distribution of the cluster assignment becomes accurate and is exploited for efficient clustering in the inference stage.

\begin{algorithm*}
\caption{We use default values of $\lambda=100$, $\eta=10$, $N_{critic}=5$, $N_{class}=4$, $N=900000$, $\alpha=0.0001$, $\beta_1=0.5$, $\beta_2=0.9$, $\tau=30000$, $\gamma=1.2$.}
\label{algo}
\begin{algorithmic}[1]
\REQUIRE The gradient penalty coefficient $\lambda$, the trade-off parameter $\eta$, the number of critic iterations $N_{critic}$ and the number of classifier iterations $N_{class}$, the number of generator iterations $N$, the batch size $n$, Adam hyperparameters $\alpha$, $\beta_1$, $\beta_2$
\REQUIRE Initial critic parameters $\theta_{d}$, initial generator parameters $\theta_{g}$, initial classifier parameters $\theta_{c}$, the number of clusters $k$
\ENSURE Classifier $C$
\WHILE{$t\leq N$ \AND $\theta_g$ has not converged}
\REPEAT
\FOR{$i=1,\ldots,n$}
\STATE Sample a real data $\textbf{x}\sim \mathbb{P}_r$, a cluster id $y\sim \mathcal{U}^{int}[1,k]$, a noise vector $\textbf{z}\sim \mathcal{N}(0,1)$, a random number $\epsilon\sim \mathcal{U}[0,1]$
\STATE $\textbf{y}\leftarrow$ one-hot coding of $y$
\STATE $\tilde{\textbf{x}}\leftarrow G_{\theta_g}(\textbf{y},\textbf{z})$
\STATE $\hat{\textbf{x}}\leftarrow \epsilon \textbf{x}+(1-\epsilon)\tilde{\textbf{x}}$
\STATE $L^{(i)}\leftarrow D_{\theta_d}(\tilde{\textbf{x}})-D_{\theta_d}(\textbf{x})+\lambda(\|\nabla_{\hat{\textbf{x}}}D_{\theta_d}(\hat{\textbf{x}})\|_2-1)^2$
\ENDFOR
\STATE $\theta_d\leftarrow Adam(\nabla_{\theta_d}\frac{1}{n}\Sigma_{i=1}^{n}L^{(i)},\theta_d,\alpha,\beta_1,\beta_2)$
\UNTIL{Reach the maximal iteration $N_{critic}$}
\REPEAT
\FOR{$i=1,\ldots,n$}
\STATE Sample a cluster id $y\sim \mathcal{U}^{int}[1,k]$, a random noise vector $\textbf{z}\sim \mathcal{N}(0,1)$
\STATE $\textbf{y}\leftarrow$ one-hot coding of $y$
\STATE $\hat{\textbf{y}}\leftarrow C_{\theta_c}(G_{\theta_g}(\textbf{y},\textbf{z}))$
\STATE $L^{(i)}\leftarrow -\Sigma_{j=1}^k \textbf{y}_j\ln \hat{\textbf{y}}_j$
\ENDFOR
\STATE $\theta_c\leftarrow Adam(\nabla_{\theta_c}\frac{1}{n}\Sigma_{i=1}^{n}L^{(i)},\theta_c,\alpha,\beta_1,\beta_2)$
\UNTIL{Reach the maximal iteration $N_{class}$}
\STATE Sample a batch of cluster ids $\{y^{(i)}\}_{i=1}^n\sim \mathcal{U}^{int}[1,k]$, a batch of random noise vectors $\{\textbf{z}^{(i)}\}_{i=1}^n\sim \mathcal{N}(0,1)$
\FORALL{$y^{(i)}$ such that $i\in[1,n]$}
\STATE $\textbf{y}^{(i)}\leftarrow$ one-hot coding of $y^{(i)}$
\STATE $\hat{\textbf{y}}^{(i)}\leftarrow C_{\theta_c}(G_{\theta_g}(\textbf{y}^{(i)},\textbf{z}^{(i)}))$
\ENDFOR
\STATE $L\leftarrow-\frac{1}{n}\Sigma_{i=1}^n( D_{\theta_d}(G_{\theta_g}(\textbf{y}^{(i)},\textbf{z}^{(i)}))+\eta\Sigma_{j=1}^k\textbf{y}^{(i)}_j\ln \hat{\textbf{y}}^{(i)}_j)$
\STATE $\theta_g\leftarrow Adam(\nabla_{\theta_g}L,\theta_g,\alpha,\beta_1,\beta_2)$
\STATE $t\leftarrow t+1$
\STATE $\eta\leftarrow\eta\gamma^{\lfloor t/\tau\rfloor}$
\ENDWHILE
\end{algorithmic}
\end{algorithm*}

\section{Experiments}\label{sec:exp}

In this section, we conduct several experiments to verify the proposed method. Specifically, we experiment on a synthetic dataset, MNIST, Fashion-MNIST, Artifact-MNIST, ORL, USPS, Cifar-10 and ImageNet-10 to examine a practical and theoretically grounded direction towards solving the deep clustering problems. As popular measures in the literature, Clustering Accuracy (ACC), Normalized Mutual Information (NMI) and Adjusted Rand Index (ARI) are employed for evaluation. The value range of ACC and NMI is [0,1], and the value range of ARI is [-1,1]. It should be noted that the effectiveness of the framework strongly relies on the capability of the generative model to produce realistic samples. Therefore, the reported results can consistently get improved as the generative model evolves, which is now prospective~\cite{DBLP:conf/iclr/MiyatoKKY18,DBLP:conf/icml/ZhangGMO19,DBLP:conf/iclr/BrockDS19,DBLP:conf/cvpr/KarrasLAHLA20}.

\subsection{Networks}
Because our main purpose is to verify the utility of integrating representation learning and clustering into a unified framework, we do not carry out exhaustive architecture and hyperparameter search in all experiments, and the architectural choice and experimental configuration are similar to~\cite{gulrajani2017improved}. In particular, the generator and discriminator inherit the network structure reported in~\cite{gulrajani2017improved}. The classifier shares a similar structure with the discriminator, but the classifier has a different output layer to produce categorical probability masses. The training dynamics between these three components have been outlined in Algorithm \ref{algo}, and will be explained in detail below.

\begin{figure}
  \centering
  \includegraphics[width=8.5cm]{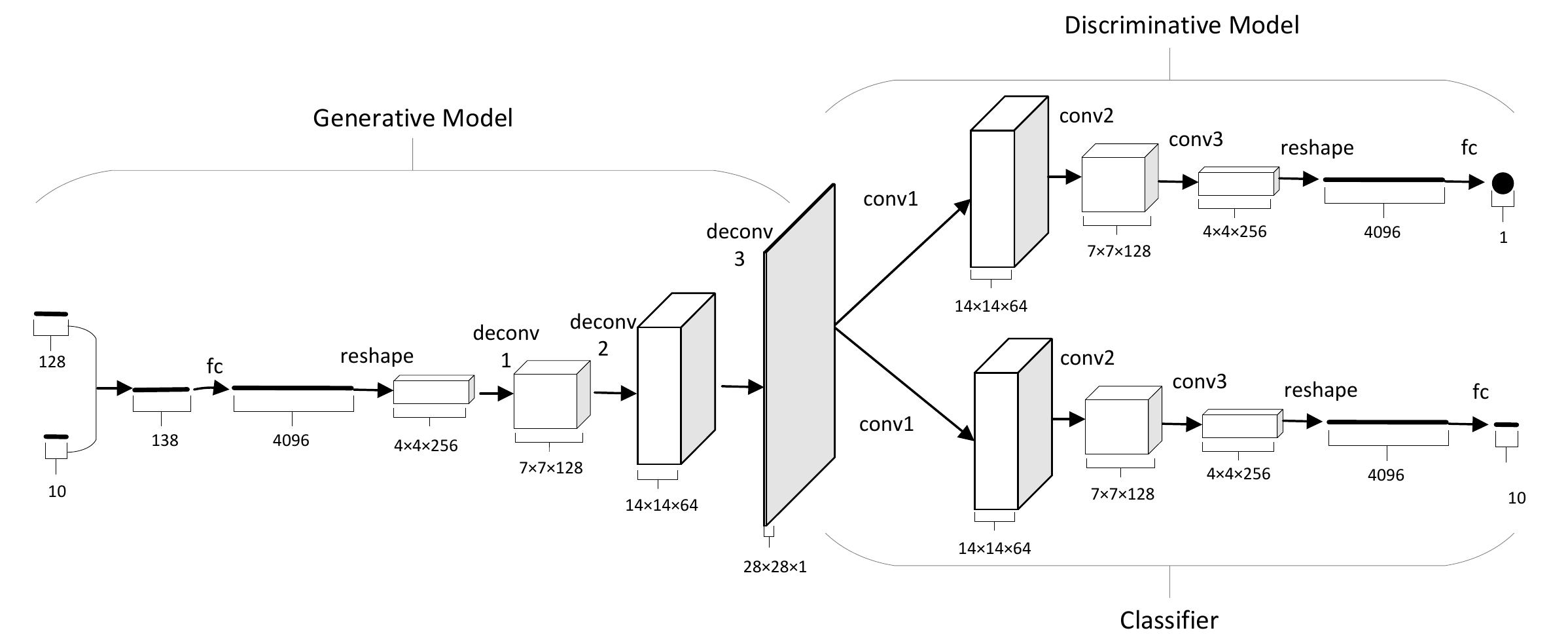}\\
  \caption{Architecture used in our experiments. This architecture is used across experiments on MNIST, Fashion-MNIST, Artifact-MNIST, ORL, USPS, Cifar-10 and ImageNet-10. For MNIST, Fashion-MNIST and Artifact-MNIST, the size of the images is 28x28, we drop one pixel horizontally and vertically after the first deconvolutional layer. In the experiments on ORL and Cifar-10, the size of the feature maps outputted by the first deconvolutional layer should be 8x8. For Cifar-10, the output of the generator should be 32x32x3, which is formalized by a deconvolutional layer with 3 kernels rather than 1 for grayscale images. As USPS variants consist of images of 16x16, we adopt a 1x1 stride in the last deconvolutional layer of the generator and the first convolutional layer of the discriminator and classifier. For ImageNet-10, we append a deconvolutional layer to the generator and a convolutional layer to the discriminator and classifier to process 64x64 images. In this figure, we just write the output of the generator as 28x28x1 for compact.}\label{fig:architecture}
\end{figure}

The networks embodying our framework are illustrated in Fig. \ref{fig:architecture}, where all the convolutional or deconvolutional layers adopt a 5x5 kernel size and a 2x2 stride. All the experiments are conducted with this architecture, and only a few modifications are made to adapt to different datasets, except for experiments on the synthetic dataset, where three fully-connected networks are employed. We have also tried to combine the discriminator and classifier into a single network with two output heads for multi-task learning. However, this strategy causes performance degradation on some data sets (i.e., according to median statistics of ten runs, at least 10\% performance degradation on MNIST and Fashion-MNIST, and completely meaningless results on Artifact-MNIST).

It is likely that even though both the discriminator and the classifier learn discriminative features of the samples, they focus on different aspects. The discriminator looks for the difference between the generated samples and real samples. As the training progresses, this difference gradually changes. On the contrary, the classifier aims to discover the accurate features for clustering. Sometimes, these two prompts may be quite different, especially in the final stage of training, at this stage, the most effective features for distinguishing the generated samples from real samples may be invalid for the on-hand clustering task. In addition, the discriminator is constrained by the gradient penalty to realize a Lipschitz function~\cite{arjovsky2017wasserstein, gulrajani2017improved}, which may impair the learning of clustering hints, whereas the separate classifier does not have to be Lipschitz.

\subsection{Experiments on synthetic dataset}
In order to study the effectiveness and characteristics of the proposed framework, we conduct experiments on a simple dataset consisting of eight isotropic Gaussian blobs of data. The centers of these Gaussian blobs are (1.414,0), (-1.414,0), (0,1.414), (0,-1.414), (1,1), (-1,1), (-1,-1), (1,-1) respectively, and the standard deviation of each blob is 0.014. The Gaussian mixture from which the samples come is figured in the supplemental material. In the experiment, we view the samples coming from the same Gaussian blob as in one cluster. The training and evaluating samples in the experiment are all randomly sampled from the Gaussian mixture. Therefore, there are indeed infinite training and evaluation samples.

In this experiment, three fully connected networks are adopted to embody the framework. The network structure of the generator acts like $x-512-512-512-2$, the network structure of the discriminator acts like $2-512-512-512-1$, and the network structure of the classifier acts like $2-512-512-512-8$. Therein $x$ denotes the number of the input variables to the generator, which varies from 8 to 16 in the experiments. As one-hot coding is applied, one cluster id corresponds to 8 input variables. The other input variables are all noise variables. We use $relu$ as activation function in each hidden layer of these networks, and use non-activation function in the output layer of the generator and discriminator. The output of the classifier is activated by $softmax$ to realize a normalized probability distribution.

A total of six experiments are performed on the synthetic dataset, where the cluster id sampled from the uniform categorical distribution on [0, 7] and different numbers (0, 1, 2, 3, 4, 8) of noise variables are used together as the input to the generator. Experimental results illustrate that when feeding 0, 1, 2 or 4 noise variables into the generator, the framework can exploit the centers of the gaussian blobs as the cue for cluster assignment. This demonstrates that, under appropriate experimental configuration, the proposed framework can exploit the dominant feature of the samples as the cue and give fascinating clustering results accordingly. However, when 3 or 8 noise variables are fed in, the generator begins to generate samples completely deviating from the true distribution, and the classifier falls into severe overfitting.

Because there are actually 3 intrinsic features (the centers of the gaussian blobs and the biases on x-axis and y-axis) that control the positions of the samples in the plane, the experimental results declare that when the number of variables used as input deviates from the true number of the intrinsic features, the performance of the framework becomes unstable. However, it is worth noting that, when the number of the input variables decreases from the actual number of the intrinsic features, the performance of the framework does not decrease as sharply as the number of the input variables increases from them on. Thus, we preferentially use fewer input variables in the framework if the true number of the intrinsic features is unknown, which is also desirable in practice for efficiency reasons. Furthermore, we can judge whether the number of the input variables exceeds the true number of the intrinsic features by plotting the learning curves. When too many variables are used as input, the Earth-Mover distance and the evaluated clustering accuracy will gradually diverge in the later stage of the training process. More details about the experimental results can be found in the supplemental material.

\subsection{Experiments on MNIST and Fashion-MNIST}\label{sec:mnist}
In academia, MNIST is a dataset widely used by members of the AI/ML/Data Science community. However, MNIST is too easy and overused. Fashion-MNIST is intended to serve as a direct drop-in replacement for the original MNIST dataset for benchmarking machine learning algorithms. Fashion-MNIST shares the same image size and structure of training and testing splits as MNIST~\cite{xiao2017fashion}. On these two datasets, we use the original training and testing splits to train and evaluate our framework and all the other comparison methods, where label information is not used during the training phase. In order to ensure the fairness of the comparison, all the comparison methods use the hyperparameters reported in the literature to retrain on the datasets. Our experiments on MNIST and Fashion-MNIST adopt the same configuration, including architecture selection and hyperparameter settings. On each dataset, we run each method ten times with the same configuration, and report the minimum, maximum, and median statistics of the three metrics. The results are summarized in Tables \ref{tab:mnist} and \ref{tab:fashionmnist}, where the results marked with $^*$ are reported in the literature.

The experimental results in Tables \ref{tab:mnist} and \ref{tab:fashionmnist} demonstrate that our method obtains state-of-the-art clustering results on the MNIST and Fashion-MNIST datasets. On MNIST, from the median statistics of the three metrics, our method is competitive with the other state-of-the-art methods. From the maximum statistics, our method is superior to all the other methods. On Fashion-MNIST, our method outperforms all the other comparison methods with a surprising advantage. The only downside of our method is that the minimum statistics of the performance on MNIST are a little bit lower compared to other state-of-the-art due to the unstable performance that has been quantitatively reflected in Fig. \ref{stability_mnist} as the considerable standard deviation. More detailed comparison of the performance of ten runs can be found in the supplemental material.

Fig. \ref{fig:learning_curves} illustrates the learning dynamics on MNIST and Fashion-MNIST. In Fig. \ref{fig:learning_curves}(a) and (e), the cross-entropy loss decreases quickly as the training begins. Correspondingly, the fake classification accuracy obtained by treating the fake labels as ground truth is rapidly improved. Later in the training process, the cross-entropy loss is kept small, and the fake classification accuracy remains close to 1. According to the derivation in Section \ref{sec:1}, since the classifier is fully optimized, the objective is actually to guide the generator to minimize the expectation of the distribution entropy of the cluster assignment. In the experiment, the expectation of the distribution entropy quickly reaches its optimum at the beginning of the training process, so the generator is actually optimized to produce realistic samples while keeping the distribution entropy to a minimum. In this case, as the quality of the generated samples improves (Earth Mover distance converges as in Fig. \ref{fig:learning_curves}(b) and (f)), the evaluated clustering accuracy will continue to increase (as shown in Fig. \ref{fig:learning_curves}(c) and (g)). Finally, when the generator produces high-quality samples, the framework obtains encouraging clustering results. In fact, the samples generated in Fig. \ref{fig:learning_curves}(d) and (h) show that the generated samples from the same cluster are similar in perception -- basically the same in digit or apparel type. This proves that the framework has discovered the digit or apparel type in the image as a clue for cluster assignment.

In the experiments, we initialize $\eta$ to 10 and then multiply it by 1.2 every 30,000 iterations. A total of 900,000 iterations of this optimization are performed. It should be noted that in Fig. \ref{fig:learning_curves}(b) and (f), the plots of the trade-off parameter are scaled by 10. In addition, the hyperparameter $\lambda$ is set to 100.

%The experimental results on MNIST and Fashion-MNIST have presented sound evidence of the effectiveness of the proposed unsupervised classification framework. The learning curves also persuade that our framework has reproduced the decision-making mechanism of humans in unsupervised classification scenario.
\begin{table*}
\caption{Performance comparison on MNIST. The best results are highlighted in \textbf{bold}. $^*$ indicates that the result is reported in literature}
\label{tab:mnist}       % Give a unique label
\begin{center}
\begin{tabular}{|c|c|c|c|c|c|c|c|c|c|}
\hline
\multirow{2}{*}{Method} & \multicolumn{3}{c|}{ACC} & \multicolumn{3}{c|}{NMI} & \multicolumn{3}{c|}{ARI} \cr\cline{2-10}
& \textit{Min} & \textit{Max} & \textit{Med} & \textit{Min} & \textit{Max} & \textit{Med} & \textit{Min} & \textit{Max} & \textit{Med} \cr
\hline\hline
NMF~\cite{cai2009locality} & - & - & $0.545^*$ & - & - & $0.608^*$ & - & - & $0.430^*$ \cr\hline
K-means~\cite{6817617} & 0.534 & 0.571 & 0.563 & 0.479 & 0.521 & 0.499 & 0.347 & 0.374 & 0.352 \cr\hline
SC~\cite{zelnik2005self} & - & - & $0.696^*$ & - & - & $0.663^*$ & - & - & $0.521^*$ \cr\hline
AC~\cite{gowda1978agglomerative} & - & - & $0.695^*$ & - & - & $0.609^*$ & - & - & $0.481^*$ \cr\hline
DeCNN~\cite{5539957} & - & - & $0.818^*$ & - & - & $0.758^*$ & - & - & $0.669^*$ \cr\hline
GAN~\cite{radford2015unsupervised} & - & - & $0.828^*$ & - & - & $0.764^*$ & - & - & $0.736^*$ \cr\hline
DDC~\cite{DBLP:journals/kbs/RenWLX20} & - & - & $0.965^*$ & - & - & $0.916^*$ & - & - & - \cr\hline
DDC-DA~\cite{DBLP:journals/kbs/RenWLX20} & - & - & $\textbf{0.970}^*$ & - & - & $0.927^*$ & - & - & - \cr\hline
DAC~\cite{Chang2017Deep} & 0.745 & 0.813 & 0.804 & 0.782 & 0.836 & 0.820 & 0.678 & 0.756 & 0.728 \cr\hline
DEC~\cite{xie2016unsupervised} & 0.862 & 0.864 & 0.864 & 0.833 & 0.835 & 0.835 & 0.797 & 0.801 & 0.800 \cr\hline
JULE~\cite{yang2016joint} & 0.948 & 0.964 & 0.960 & 0.901 & 0.926 & 0.912 & 0.913 & 0.927 & 0.922 \cr\hline
SpectralNet~\cite{shaham2018spectralnet} & \textbf{0.967} & 0.971 & 0.969 & 0.920 & 0.924 & 0.921 & \textbf{0.931} & 0.934 & \textbf{0.933} \cr\hline
DCCM~\cite{DBLP:conf/iccv/WuLWQLLZ19} & 0.641 & 0.921 & 0.780 & 0.651 & 0.905 & 0.785 & 0.499 & 0.815 & 0.650 \cr\hline
Ours & 0.915 & \textbf{0.984} & 0.958 & \textbf{0.922} & \textbf{0.978} & \textbf{0.944} & 0.855 & \textbf{0.951} & 0.912 \cr
\hline
\end{tabular}
\end{center}
\end{table*}

\begin{table*}
\caption{Performance comparison on Fashion-MNIST. The best results are highlighted in \textbf{bold}. The improvement of our method has built a large margin on three metrics regardless of the used statistics}
\label{tab:fashionmnist}       % Give a unique label
\begin{center}
\begin{tabular}{|c|c|c|c|c|c|c|c|c|c|}
\hline
\multirow{2}{*}{Method} & \multicolumn{3}{c|}{ACC} & \multicolumn{3}{c|}{NMI} & \multicolumn{3}{c|}{ARI} \cr\cline{2-10}
& \textit{Min} & \textit{Max} & \textit{Med} & \textit{Min} & \textit{Max} & \textit{Med} & \textit{Min} & \textit{Max} & \textit{Med} \cr
\hline\hline
DDC~\cite{DBLP:journals/kbs/RenWLX20} & - & - & $0.619^*$ & - & - & $0.682^*$ & - & - & - \cr\hline
DDC-DA~\cite{DBLP:journals/kbs/RenWLX20} & - & - & $0.609^*$ & - & - & $0.661^*$ & - & - & - \cr\hline
K-means~\cite{6817617} & 0.254 & 0.354 & 0.331 & 0.172 & 0.307 & 0.256 & 0.181 & 0.271 & 0.249 \cr\hline
DEC~\cite{xie2016unsupervised} & 0.469 & 0.478 & 0.477 & 0.492 & 0.504 & 0.501 & 0.320 & 0.331 & 0.330 \cr\hline
SpectralNet~\cite{shaham2018spectralnet} & 0.488 & 0.523 & 0.505 & 0.519 & 0.529 & 0.523 & 0.329 & 0.347 & 0.337 \cr\hline
JULE~\cite{yang2016joint} & 0.423 & 0.505 & 0.486 & 0.594 & 0.652 & 0.639 & 0.342 & 0.421 & 0.390 \cr\hline
DAC~\cite{Chang2017Deep} & 0.435 & 0.591 & 0.531 & 0.487 & 0.584 & 0.552 & 0.371 & 0.459 & 0.414 \cr\hline
DCCM~\cite{DBLP:conf/iccv/WuLWQLLZ19} & 0.406 & 0.593 & 0.544 & 0.315 & 0.515 & 0.449 & 0.301 & 0.416 & 0.375 \cr\hline
Ours & \textbf{0.685} & \textbf{0.754} & \textbf{0.721} & \textbf{0.689} & \textbf{0.749} & \textbf{0.719} & \textbf{0.524} & \textbf{0.589} & \textbf{0.555} \cr
\hline
\end{tabular}
\end{center}
\end{table*}

\begin{figure*}[tb]
\centering
\begin{minipage}{1\linewidth}
\begin{minipage}{0.24\linewidth}
  \centerline{\includegraphics[width=1.767in]{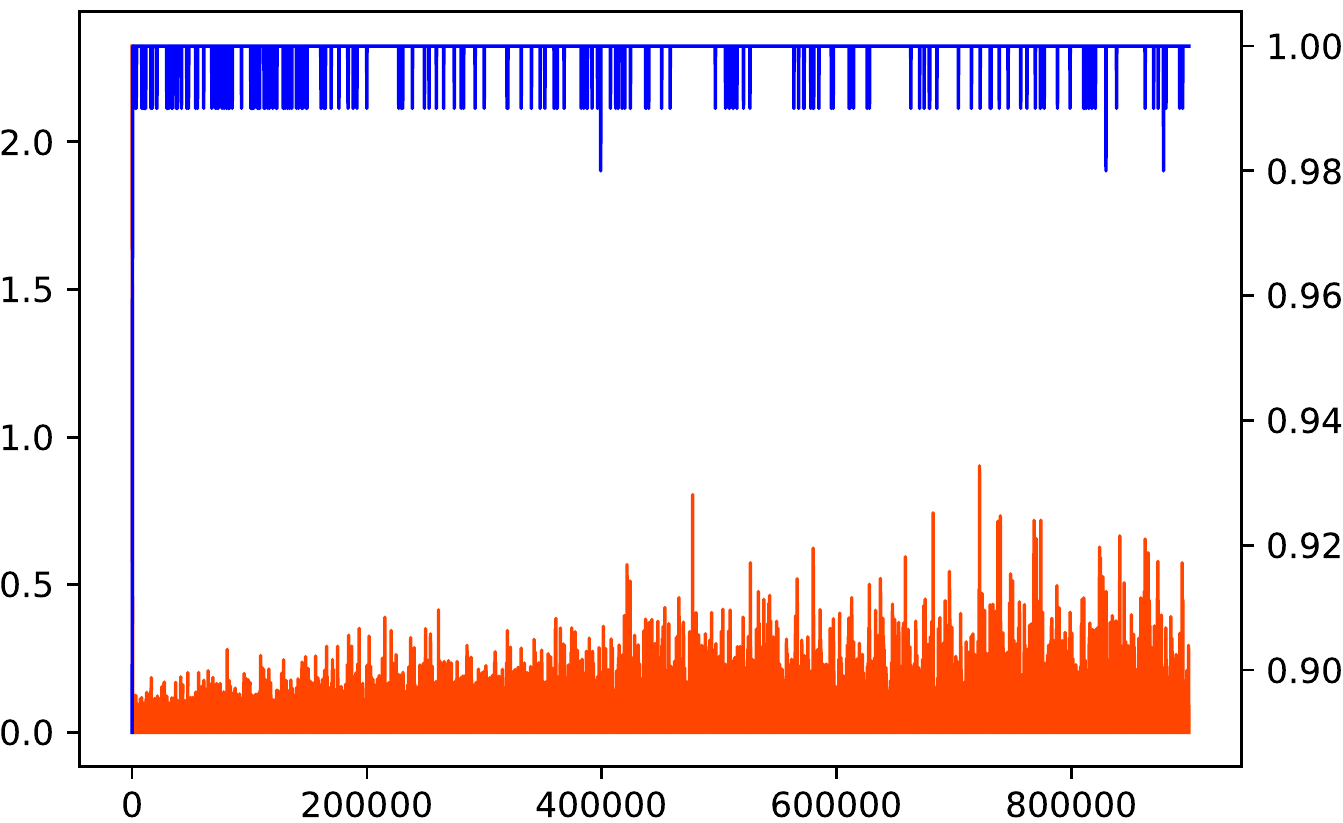}}
  \centerline{(a)}
\end{minipage}
\hfill
\begin{minipage}{0.245\linewidth}
  \centerline{\includegraphics[width=1.78in]{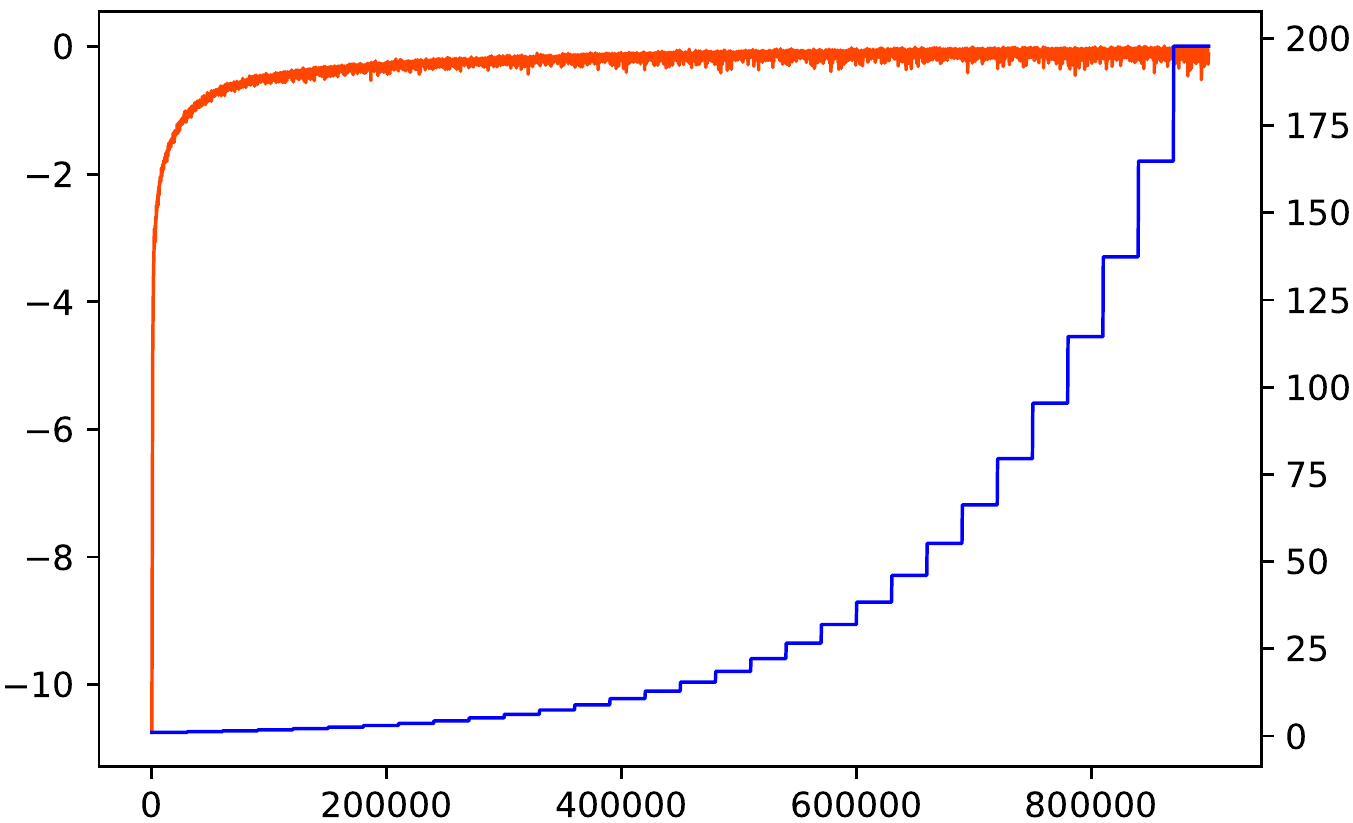}}
  \centerline{(b)}
\end{minipage}
\hfill
\begin{minipage}{0.23\linewidth}
  \centerline{\includegraphics[width=1.65in]{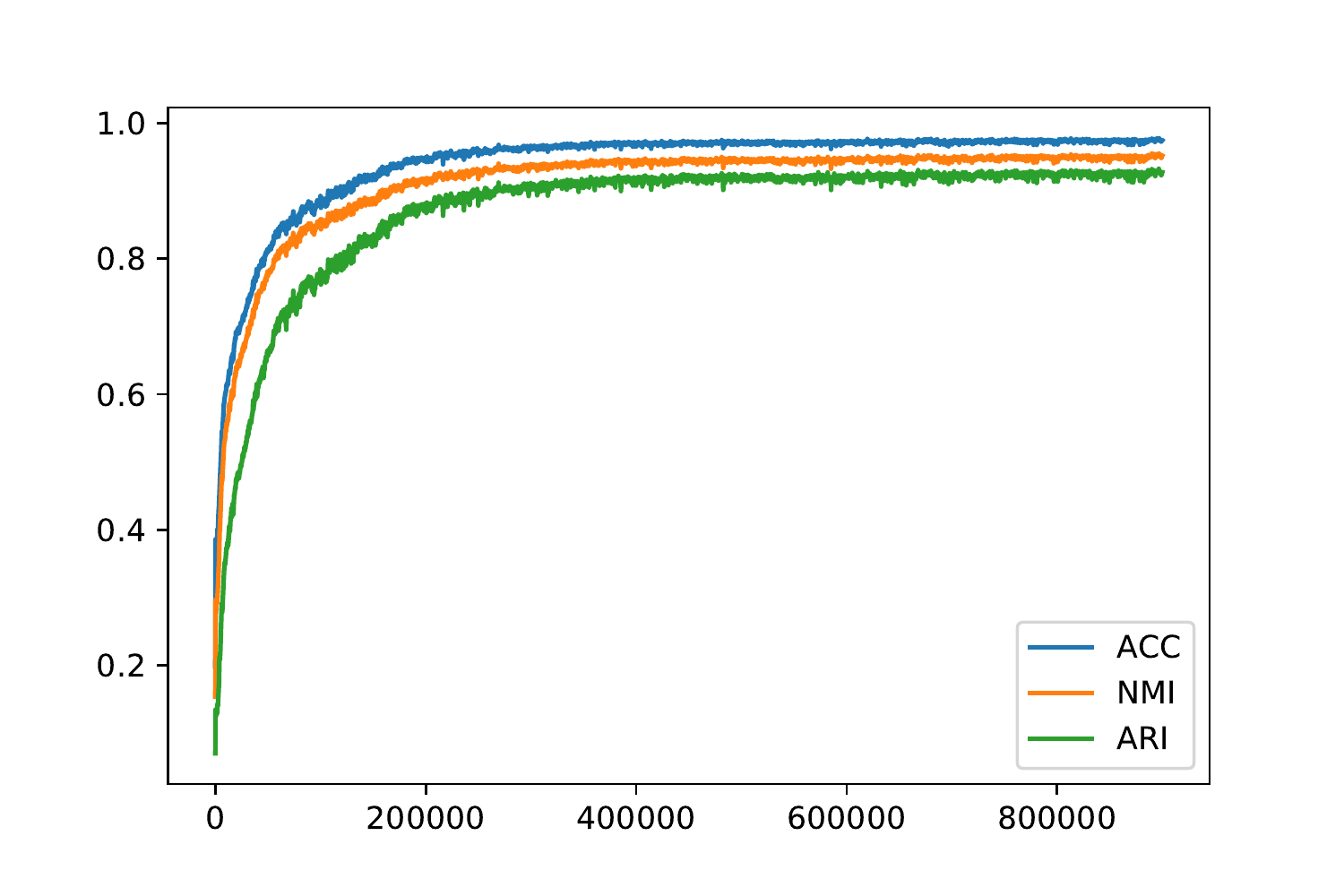}}
  \centerline{(c)}
\end{minipage}
\hfill
\begin{minipage}{0.245\linewidth}
  \vspace{4pt}
  \centerline{\includegraphics[width=1.78in]{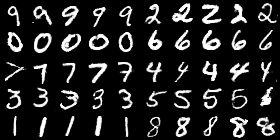}}
  \vspace{8pt}
  \centerline{(d)}
\end{minipage}
\end{minipage}
\vfill
\begin{minipage}{1\linewidth}
\begin{minipage}{0.24\linewidth}
  \centerline{\includegraphics[width=1.767in]{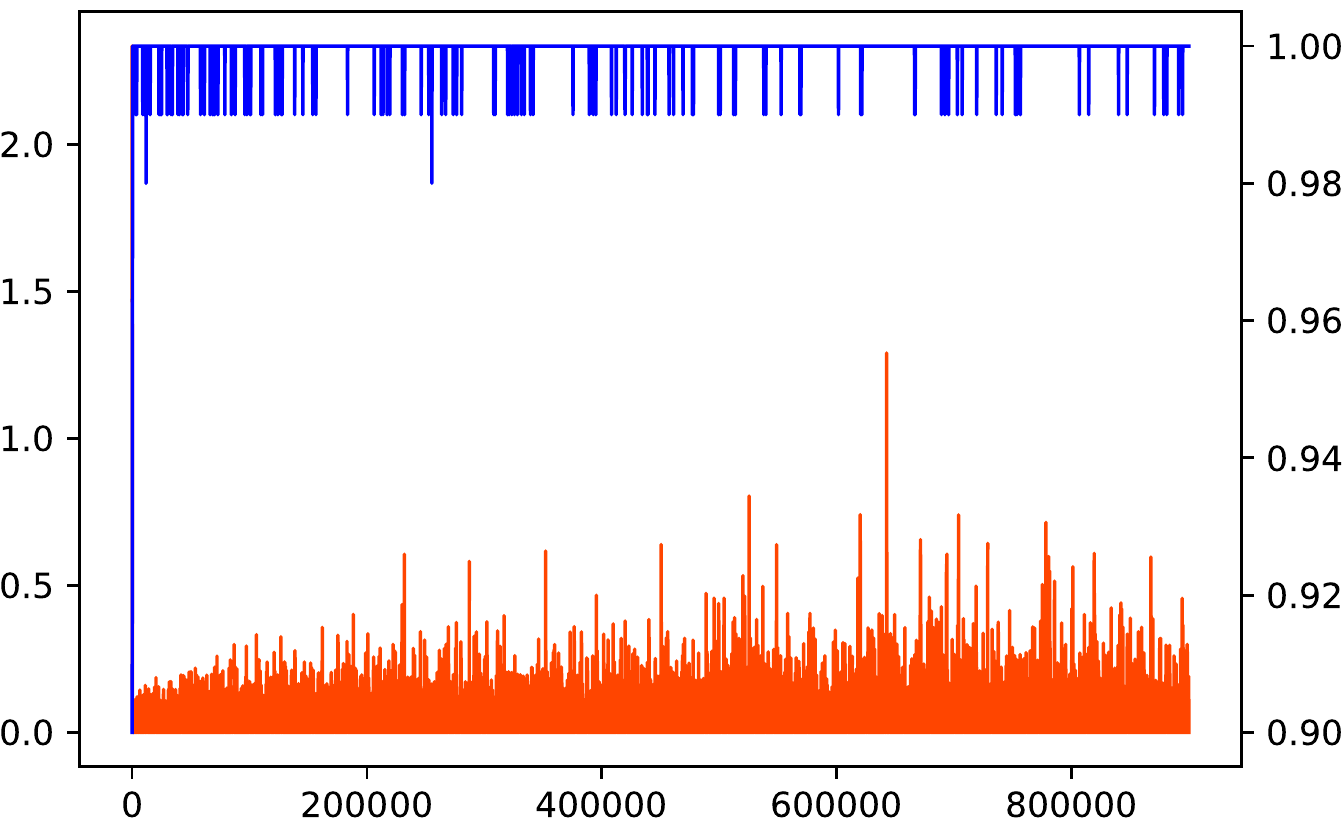}}
  \centerline{(e)}
\end{minipage}
\hfill
\begin{minipage}{0.245\linewidth}
  \centerline{\includegraphics[width=1.78in]{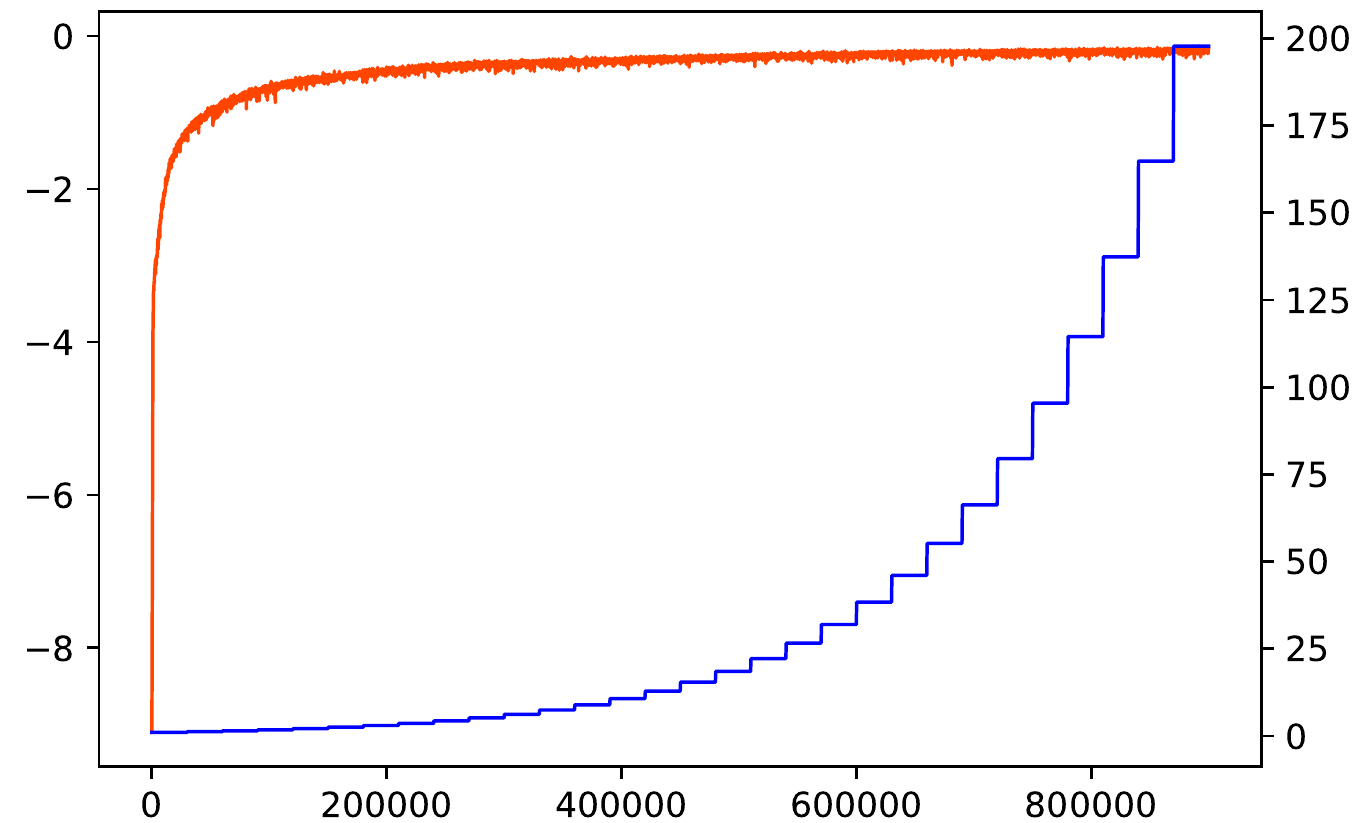}}
  \centerline{(f)}
\end{minipage}
\hfill
\begin{minipage}{0.23\linewidth}
  \centerline{\includegraphics[width=1.65in]{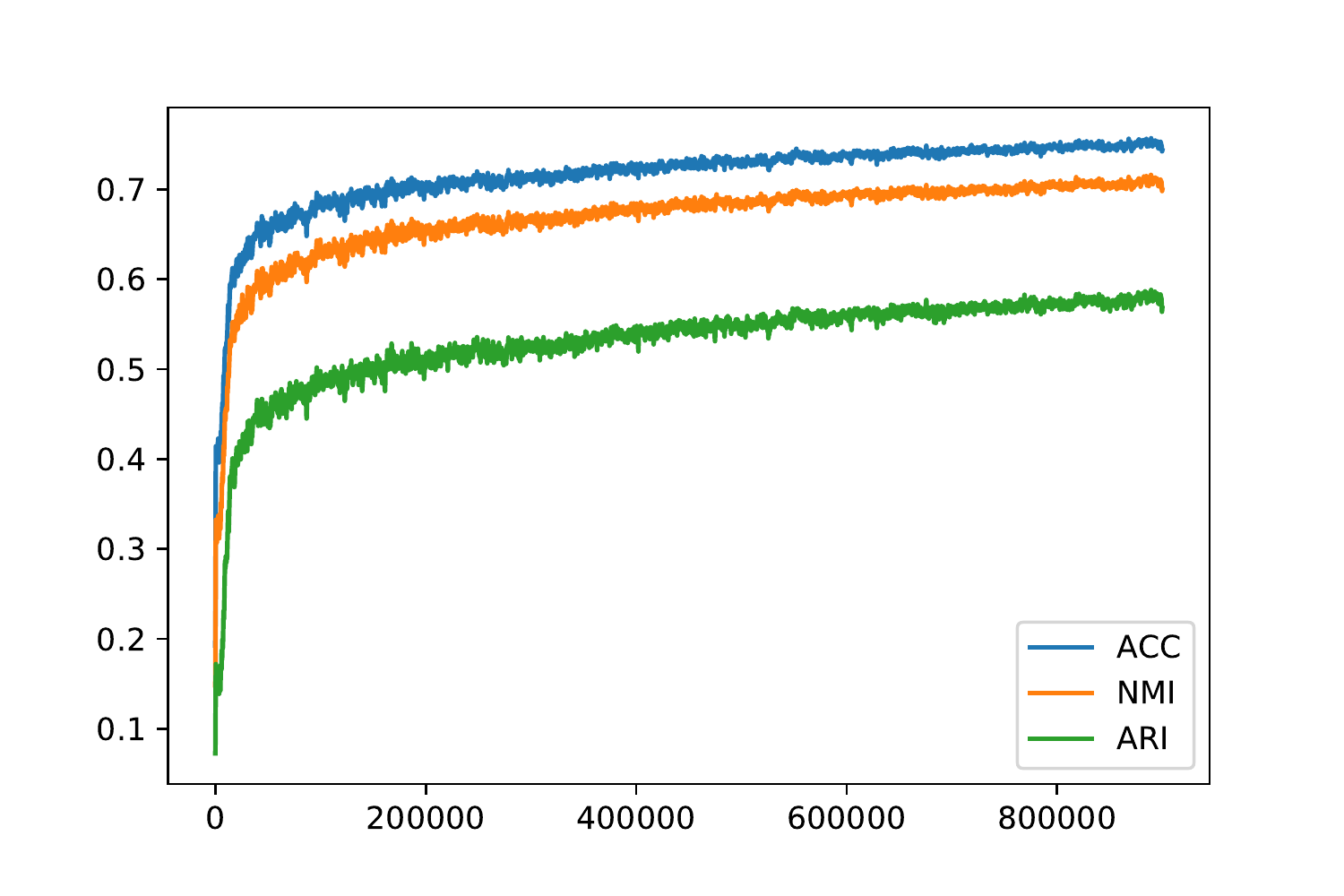}}
  \centerline{(g)}
\end{minipage}
\hfill
\begin{minipage}{0.245\linewidth}
  \vspace{4pt}
  \centerline{\includegraphics[width=1.78in]{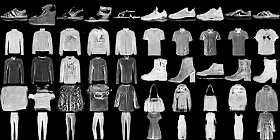}}
  \vspace{8pt}
  \centerline{(h)}
\end{minipage}
\end{minipage}
\caption{Learning curves on MNIST and Fashion-MNIST. The subfigures in the first line depicts the learning dynamics on MNIST, and the subfigures in the second line depicts the learning dynamics on Fashion-MNIST. (a) and (e) illustrate the changes of the expectation of distribution entropy of the cluster assignment (orange, ticks on the left axis), and the fake classification accuracy (blue, ticks on the right axis) during training. (b) and (f) describe the Earth Mover distance evaluated on the test data (orange, ticks on the left axis), and the trade-off parameter $\eta$ (blue, ticks on the right axis) at each iteration. (c) and (g) display the evolution of the ACC, NMI and ARI metrics evaluated on the test data at each iteration. Finally, the samples generated at the last optimization iteration are given in (d) and (h). The generated samples are arranged according to the cluster id. Specifically, the first five or last five samples of each line in (d) and (h) are generated from the same cluster id. A total of 900,000 iterations of this optimization are performed.}
\label{fig:learning_curves}
\end{figure*}

%\begin{figure}
%  \centering
%  % Requires \usepackage{graphicx}
%  \includegraphics[width=7cm]{mnist.pdf}\\
%  \caption{Learning curves on MNIST. The first row depicts the variance of classifying cross entropy loss (expectation of entropy of the posterior distribution of class label indeed), wasserstein distance, and fake classifying accuracy during training. The second row depicts the dynamics of ACC, NMI, and ARI evaluated on test data. The third row depicts the wasserstein distance evaluated on test data, the variance of the trade-off parameter $\eta$ in the training process, and the generated samples in the last optimization iteration. The generated samples are arranged according to the class labels. The first five or last five samples of each row are generated from the same class label. The optimization is performed by 900000 iterations in total.}\label{fig:mnist}
%\end{figure}

%\begin{figure}
%  \centering
%  % Requires \usepackage{graphicx}
%  \includegraphics[width=7cm]{fashion-mnist.pdf}\\
%  \caption{Learning curves on Fashion-MNIST. All the components are the same as that in Fig.\ref{fig:mnist}. The experiment on Fashion-MNIST adopts the same configuration as experiment on MNIST, including architecture and hyperparameters.}\label{fig:fashionmnist}
%\end{figure}

\subsection{Learning a precise feature as the cue for clustering}
We have argued in Section \ref{sec:method} that our objective is to learn an exactly matched characteristic as the cue for cluster assignment, which differs the proposed framework from previous methods. However, the above experiments cannot provide proof of this statement, because the clustering tasks are still focused on finding the most dominant feature (centers of the gaussian blobs in synthetic dataset, digit and apparel types in MNIST and Fashion-MNIST) of the data points, and performing grouping accordingly. In this section, we plan to empirically demonstrate the ability of the proposed framework to learn a precise feature as the cue for cluster assignment.
\subsubsection{Experiments on Artifact-MNIST} To this end, we make an artifact version of the original MNIST dataset and name it as Artifact-MNIST. Specifically, we randomly set the first pixel of each image in the MNIST dataset to 0.1, 0.5, and 0.9 with the same probability. Since the artifact is so subtle, it must not be the dominant feature of the samples, and can be regarded as an analogy of chirography. In addition, because of the sample-level and independent (of other features) nature, the artifact is one of the intrinsic characteristics of these samples. In this experiment, we want to investigate whether our method can find such delicate artifacts as clues for cluster assignment when three categories are specified for the clustering task.

We run our method and all the comparison methods ten times on Artifact-MNIST. The experimental results are listed in Table \ref{tab:artifactmnist}, where the zero values of NMI and ARI represent meaningless (totally random) cluster assignment. A more detailed comparison of ten runs is given in the supplementary material. Although our method does not always provide ideal results, all the other clustering algorithms can not accomplish this task. The reason for the failure cases of our method may be that the dataset contains other intrinsic features that are also ternary. The intelligent agent tries to capture an intrinsic feature to solve this three-category clustering problem, but it will arbitrarily select one from multiple candidates, and does not always encounter the desired one. However, the consistent success results (over 80\% of the time) indicate that our framework can indeed solve such general-purpose clustering problems. In practice, a small evaluation set is needed to check whether or not the intelligent agent has captured the desirable feature as the cue for cluster assignment.

\begin{table*}
\caption{Performance comparison on Artifact-MNIST. The best results are highlighted in \textbf{bold}. The zero values of NMI and ARI indicate completely random cluster assignment}
\label{tab:artifactmnist}       % Give a unique label
\begin{center}
\begin{tabular}{|c|c|c|c|c|c|c|c|c|c|}
\hline
\multirow{2}{*}{Method} & \multicolumn{3}{c|}{ACC} & \multicolumn{3}{c|}{NMI} & \multicolumn{3}{c|}{ARI} \cr\cline{2-10}
& \textit{Min} & \textit{Max} & \textit{Med} & \textit{Min} & \textit{Max} & \textit{Med} & \textit{Min} & \textit{Max} & \textit{Med} \cr
\hline\hline
K-means~\cite{6817617} & 0.347 & 0.352 & 0.350 & 0.000 & 0.000 & 0.000 & 0.000 & 0.000 & 0.000 \cr\hline
DEC~\cite{xie2016unsupervised} & 0.335 & 0.337 & 0.336 & 0.000 & 0.000 & 0.000 & 0.000 & 0.000 & 0.000 \cr\hline
SpectralNet~\cite{shaham2018spectralnet} & 0.338 & 0.342 & 0.340 & 0.000 & 0.000 & 0.000 & 0.000 & 0.000 & 0.000 \cr\hline
JULE~\cite{yang2016joint} & 0.345 & 0.362 & 0.359 & 0.000 & 0.000 & 0.000 & 0.000 & 0.000 & 0.000 \cr\hline
DAC~\cite{Chang2017Deep} & 0.348 & 0.379 & 0.368 & 0.000 & 0.000 & 0.000 & 0.000 & 0.000 & 0.000 \cr\hline
DCCM~\cite{DBLP:conf/iccv/WuLWQLLZ19} & 0.322 & 0.349 & 0.332 & 0.000 & 0.000 & 0.000 & 0.000 & 0.000 & 0.000 \cr\hline
Ours & \textbf{0.467} & \textbf{1.000} & \textbf{1.000} & \textbf{0.051} & \textbf{1.000} & \textbf{1.000} & \textbf{0.036} & \textbf{1.000} & \textbf{1.000} \cr
\hline
\end{tabular}
\end{center}
\end{table*}

\subsubsection{Experiments on the ORL dataset}
\begin{table*}
\caption{Performance of various clustering methods on ORL. The best results are highlighted in \textbf{bold}. In each cell, the left value is examined by viewing identity as ground truth, and the right value is examined by viewing gender as ground truth}
\label{tab:orl}       % Give a unique label
\begin{center}
\setlength{\tabcolsep}{1mm}{
\begin{tabular}{|c|cc|cc|cc|cc|cc|cc|cc|cc|cc|}
\hline
\multirow{2}{*}{Method} & \multicolumn{6}{c|}{ACC} & \multicolumn{6}{c|}{NMI} & \multicolumn{6}{c|}{ARI} \cr\cline{2-19}
& \multicolumn{2}{c|}{\textit{Min}} & \multicolumn{2}{c|}{\textit{Max}} & \multicolumn{2}{c|}{\textit{Med}} & \multicolumn{2}{c|}{\textit{Min}} & \multicolumn{2}{c|}{\textit{Max}} & \multicolumn{2}{c|}{\textit{Med}} & \multicolumn{2}{c|}{\textit{Min}} & \multicolumn{2}{c|}{\textit{Max}} & \multicolumn{2}{c|}{\textit{Med}} \cr
\hline\hline
DAC\cite{Chang2017Deep} & 0.098 & 0.265 & 0.143 & 0.690 & 0.133 & 0.404 & 0.286 & 0.027 & 0.379 & 0.082 & 0.344 & 0.049 & 0.009 & -0.020 & 0.052 & 0.150 & 0.040 & 0.008 \cr\hline
K-means\cite{6817617} & 0.715 & 0.900 & 0.800 & 0.900 & 0.751 & 0.900 & 0.842 & 0.018 & 0.885 & 0.020 & 0.864 & 0.019 & 0.570 & -0.009 & 0.673 & -0.005 & 0.621 & -0.007 \cr\hline
DEC\cite{xie2016unsupervised} & 0.025 & 0.503 & 0.395 & 0.900 & 0.191 & 0.569 & 0.000 & 0.000 & 0.619 & 0.058 & 0.422 & 0.006 & 0.000 & -0.039 & 0.203 & 0.036 & 0.073 & 0.000 \cr\hline
SpectralNet\cite{shaham2018spectralnet} & 0.025 & 0.888 & 0.448 & 0.892 & 0.025 & 0.890 & 0.000 & 0.007 & 0.670 & 0.009 & 0.000 & 0.008 & 0.000 & -0.020 & 0.293 & -0.012 & 0.000 & -0.016 \cr\hline
JULE\cite{yang2016joint} & 0.560 & 0.900 & 0.625 & 0.900 & 0.597 & 0.900 & 0.758 & 0.006 & 0.805 & 0.093 & 0.786 & 0.030 & 0.371 & -0.084 & 0.480 & 0.099 & 0.428 & -0.025 \cr\hline
\tiny{$S^2ConvSCN-l_2$}\cite{DBLP:conf/cvpr/ZhangLYQZ0L19} & - & - & - & - & $0.888^*$ & - & - & - & - & - & - & - & - & - & - & - & - & - \cr\hline
\tiny{$S^2ConvSCN-l_1$}\cite{DBLP:conf/cvpr/ZhangLYQZ0L19} & - & - & - & - & $0.895^*$ & - & - & - & - & - & - & - & - & - & - & - & - & - \cr\hline
DCCM~\cite{DBLP:conf/iccv/WuLWQLLZ19} & 0.735 & 0.502 & 0.825 & 0.610 & 0.775 & 0.540 & \textbf{0.883} & 0.000 & \textbf{0.921} & 0.163 & \textbf{0.902} & 0.025 & 0.651 & -0.016 & 0.764 & 0.047 & 0.705 & 0.002 \cr\hline
Ours & \textbf{0.893} & \textbf{0.973} & \textbf{0.922} & \textbf{0.980} & \textbf{0.910} & \textbf{0.978} & 0.875 & \textbf{0.615} & 0.902 & \textbf{0.840} & 0.893 & \textbf{0.688} & \textbf{0.875} & \textbf{0.615} & \textbf{0.902} & \textbf{0.840} & \textbf{0.893} & \textbf{0.688} \cr
\hline
\end{tabular}}
\end{center}
\end{table*}

ORL~\cite{DBLP:conf/wacv/SamariaH94} is a widely used dataset in the context of face recognition~\cite{DBLP:conf/cvpr/ZhangLYQZ0L19, 9210834}. Images in the dataset are taken under varying illumination conditions, facial expressions, and facial occlusions (with or without glasses). The ORL dataset consists of 400 images, 10 each of 40 different subjects. There are 4 female and 36 male subjects in the dataset. The ORL dataset can be used for verifying the capability of finding precise feature as the cue for cluster assignment. In the dataset, clustering can be performed according to identities or genders of the facial images.

In this experiment, each facial image is resized to 32x32 pixels. Since there are only 400 samples in the dataset, we optimize the generator for 30,000 iterations. Correspondingly, we initialize $\eta$ to 10 and then multiply it by 1.2 every 1,000 iterations (reduced in proportion to the number of the optimization iterations). All the comparison methods are also trained on the dataset using the default configuration on MNIST. We run each method for ten times, and the statistics are listed in Table \ref{tab:orl}. In each cell of Table \ref{tab:orl}, the left value is obtained by performing clustering according to identity, and the right value is obtained by performing clustering according to gender. When we use the framework to perform clustering according to gender, we provide the framework not only the categorical information, but also the distribution of the clusters, as we know that there are only four females. In spite of this, when we regard gender as the ground truth, the clustering performance is still less satisfying, because there is a hard subject in the dataset (the 12th subject), for which it is difficult for even humans to judge his gender.

It should be noted that clustering according to gender is a binary clustering task with extremely unbalanced distribution, so completely random assignment means an ACC around 0.5, and putting all samples into one cluster corresponds to an ACC of 0.9. Therefore, when performing clustering according to gender, the indicators NMI and ARI are of more reference values than ACC. With reference to Table \ref{tab:orl}, it can be found that, except our method, all methods fail to cluster by gender. This experiment on a real-world face recognition dataset proves that our method does possess the ability to finding precise cue (identity or gender) of the samples according to the specified number of the categories for cluster assignment.

\subsection{Stability analysis}
\begin{figure}[!ht]
\centering
\subfloat[]{\includegraphics[width=4.3cm]{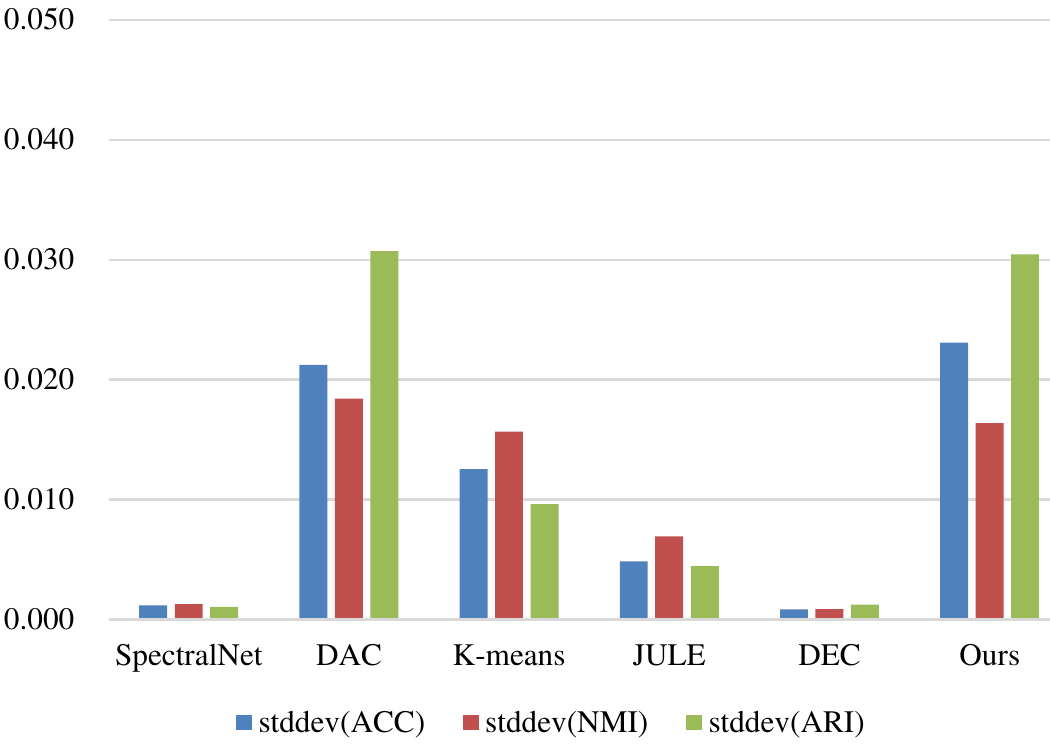}%
\label{stability_mnist}}
\hfil
\subfloat[]{\includegraphics[width=4.3cm]{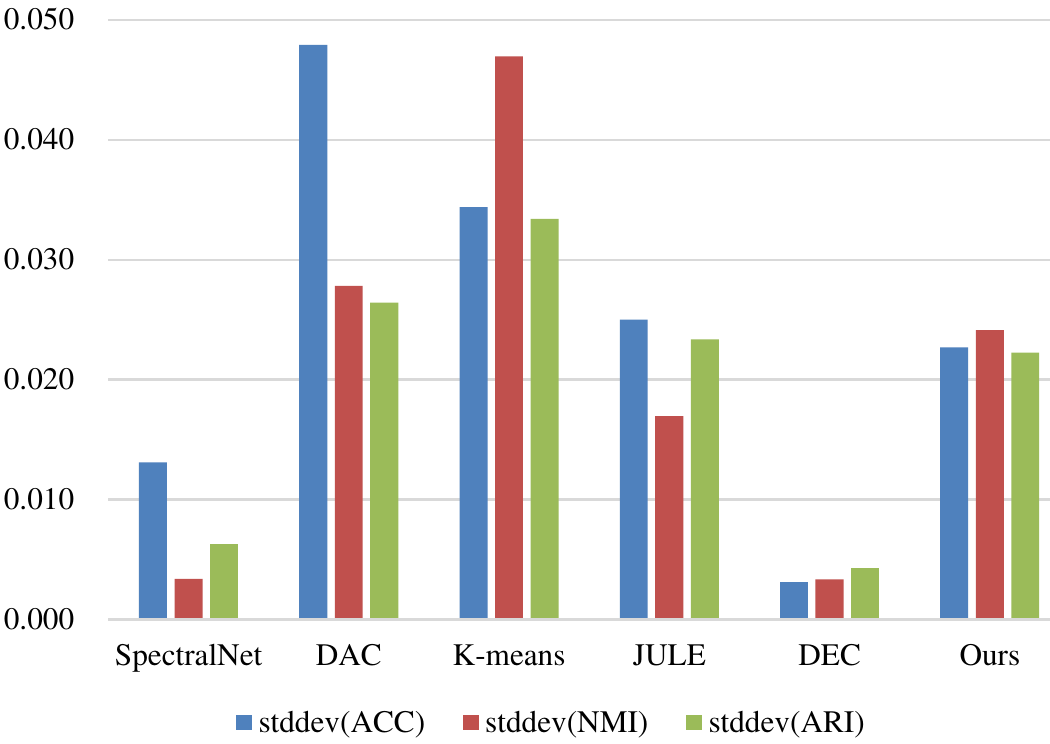}%
\label{stability_fashion}}
\caption{Comparison of the stability of the performance of different methods. (a) and (b) illustrate the results evaluated on MNIST and Fashion-MNIST respectively. In (a) and (b), the standard deviation of each indicator (ACC, NMI, and ARI) evaluated in ten runs is calculated and plotted.}
\label{fig:stability_comparison}
\end{figure}

Stability is one of many indicators to assess the quality of an algorithm. In order to compare the stability of the performance of our method and the other methods, we calculated the standard deviations of the three metrics in ten runs for all the algorithms. Finally, the results evaluated on MNIST and Fashion-MNIST are shown in Fig. \ref{fig:stability_comparison}. It can be seen that on MNIST, our method is very unstable compared to SpectralNet~\cite{shaham2018spectralnet} and DEC~\cite{xie2016unsupervised}. On Fashion-MNIST, the performance of SpectralNet~\cite{shaham2018spectralnet} and DEC~\cite{xie2016unsupervised} began to oscillate, but these two methods still perform more stable than our method.

We have tried to provide fixed seeds to the random number generators of Python and Tensorflow (we use Tensorflow for implementation). However, the resulted clustering performance still fluctuates. We attribute the instability of the clustering performance to the random behavior of the cuDNN library employed by Tensorflow. In any case, our method is sensitive to the initialization of the network parameters. This is the weakness of our method. We hope to make the proposed method more robust to random initialization in the future.

\subsection{Dealing with non-uniform distributions}
In Section \ref{sec:1}, we adopt a uniform prior for the marginal distribution of the clusters to realize the first constraint of Assumption \ref{ass:1}, which limits the application of our method to problems where samples are evenly distributed across clusters. To concretely see how the distribution of clusters affects the clustering performance, we construct five variants of the USPS dataset and conduct experiments on these variants. USPS is a handwritten digit dataset, which consists of 7,291 training and 2,007 test images. The images in USPS are of 16x16 grayscale pixels. The samples in USPS are unevenly distributed across 10 classes, with the largest class owning 1,553 samples, and the smallest class owning 708 samples. In this experiment, we reproduce the experimental configuration of Section \ref{sec:mnist}, including architecture and hyperparameters.

To quantitatively evaluate how the cluster distribution affects the performance of our method, we defined a metric to measure the uniformity of a dataset:
\begin{equation}\label{eq:uniformness}
  UI(X)=\frac{-\sum_{i=1}^n p_i\ln p_i}{\ln n},
\end{equation}
where $UI$ is the defined metric, $X$ is a dataset, $p_i$ denotes the ratio of the samples of the $ith$ class in all the samples, and $n$ denotes the number of classes. The more evenly distributed the classes are, the greater the value of the metric will be. The maximum value of $UI$ is one which indicates that the samples in the dataset are evenly distributed across all the classes, and the minimum value of $UI$ is zero which indicates that the samples in the dataset all come from one class.

\begin{figure}[!ht]
\centering
\subfloat[]{\includegraphics[width=4.3cm]{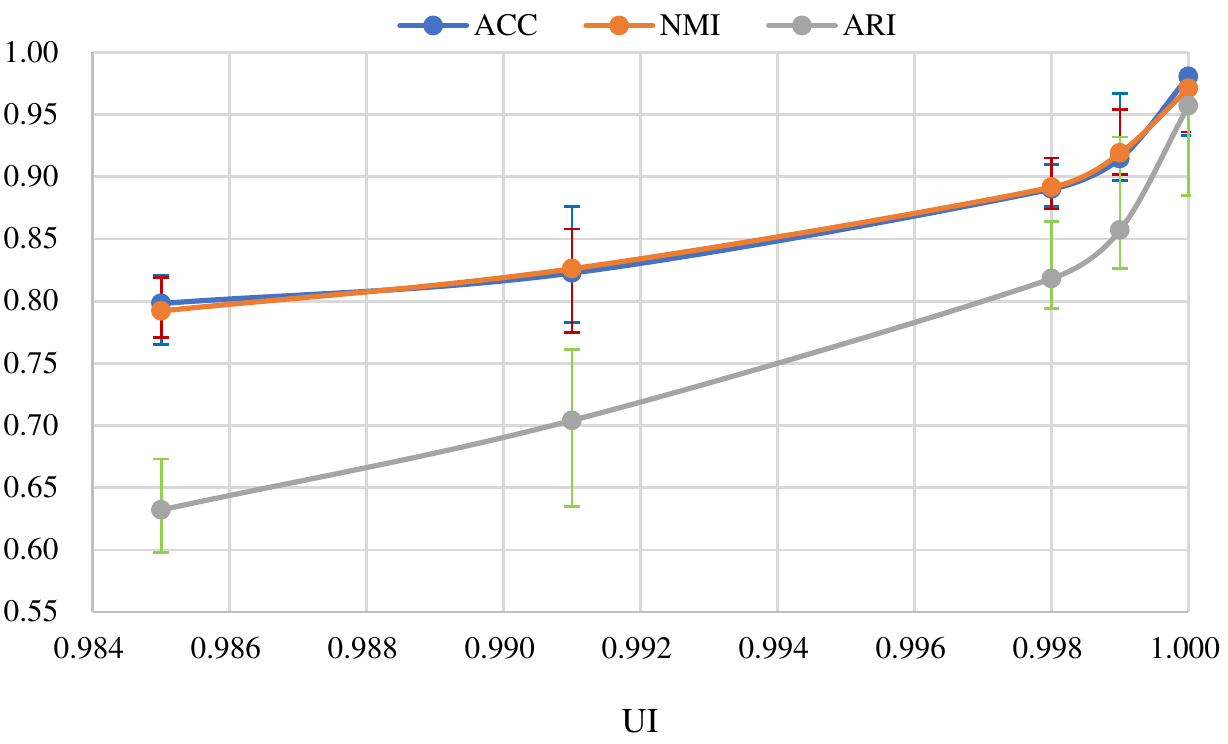}%
\label{fig:usps}}
\hfil
\subfloat[]{\includegraphics[width=4.3cm]{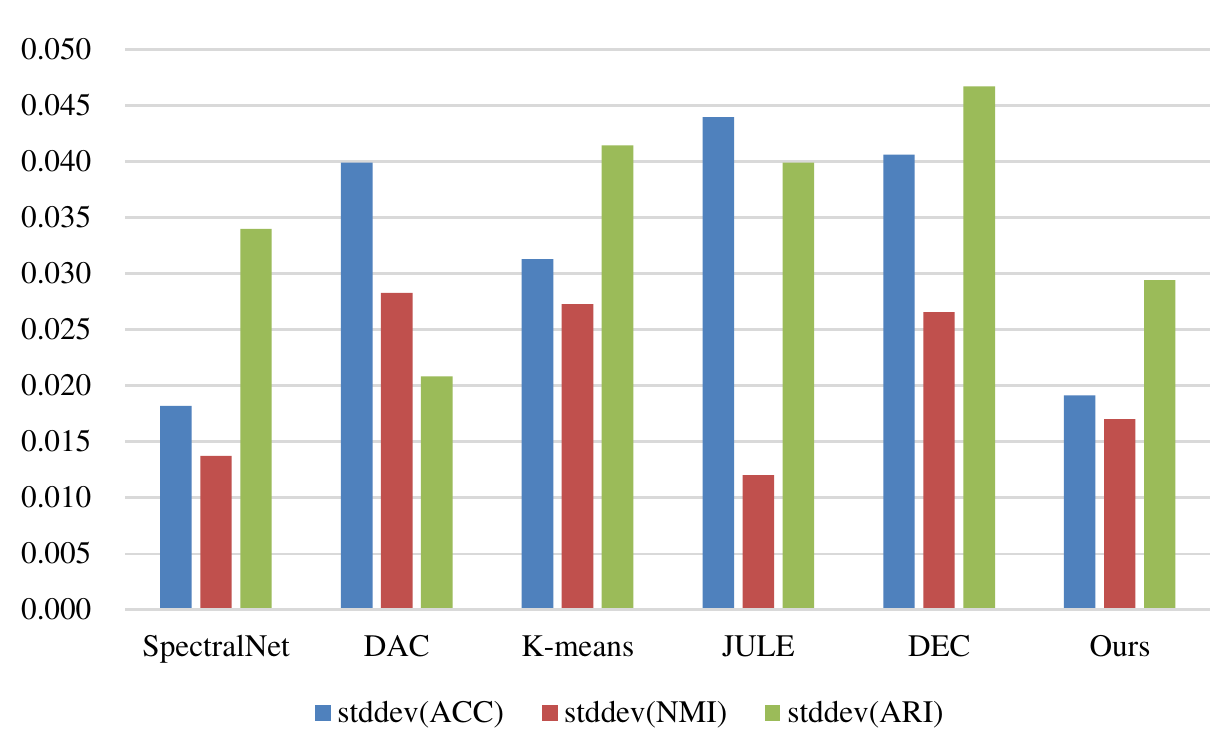}%
\label{fig:stability_usps700}}
\caption{A sketch of how the clustering performance varies with the uniformity index of the dataset and the comparison of the stability of different methods. (a) illustrates that as the UI value increases, the clustering performance also gets improved. The lower boundary of the error line represents the minimum of the performance in ten runs, and the upper boundary represents the maximum of the performance in ten runs. (b) illustrates the comparison of the stability of different methods on USPS700. The standard deviation of the performance (ACC, NMI, and ARI) of each method in ten runs is plotted.}
\label{fig:usps_result}
\end{figure}

Please note that in this experiment, we specify 10 categories for the clustering task, so the class distribution of the dataset is also the cluster distribution. In addition, the annotations are only necessary for theoretical analysis of clustering performance but not in practical applications. To create variants of the USPS dataset, we iteratively reduce the largest classes to the second largest class by removing samples. In particular, we only select an integer multiple of 100 samples from each class. Finally, we obtained five datasets with different UI values. Detailed information about these datasets can be found in the supplementary material. We evaluate our method on each of these datasets and calculate the ACC, NMI and ARI metrics. Fig. \ref{fig:usps_result}(a) illustrates how the clustering performance is affected by the UI value. It can be seen that, as the UI value increases, the performance of our method keeps improving. The reported results are median statistics of ten runs. Here, we use the same samples for training and evaluation, with label information only used in the evaluation phase.

Finally, we make a comparison between our method and the other methods on USPS700. The performance comparison is summarized in Table \ref{tab:usps700}, and the stability comparison is illustrated in Fig. \ref{fig:usps_result}(b). We use the experimental configuration reported in the literature to run each comparison method. For methods that did not carry out experiments on the USPS dataset, we resize the images to 28x28 and use the MNIST's configuration for the experiment. It can be seen that on USPS700, our method performs much better and more stable than all the comparison methods. The stable performance is really a surprise as we haven't expected it since we found the instability of our method in Section \ref{sec:mnist}. This pronounces that the stability of performance depends on specific datasets.

%\begin{figure}
%  \centering
%  % Requires \usepackage{graphicx}
%  \includegraphics[width=7cm]{USPS.pdf}\\
%  \caption{Clustering performance on variants of USPS dataset. As the UI value increases, the clustering performance of our method keeps improving. The low boundary of the error line indicates the minimum of the performance in ten runs, and the high boundary is the maximum of the performance in ten runs. The ACC curve and the NMI curve approximately overlap in this figure.}\label{fig:usps}
%\end{figure}

%\begin{table}\label{tab:datasets}
%\caption{Variants of USPS dataset.}
%\renewcommand\arraystretch{1.2}
%\begin{center}
%\begin{tabular}{|c|c|c|c|c|c|c|}
%  \hline
%  % after \\: \hline or \cline{col1-col2} \cline{col3-col4} ...
%  \multicolumn{2}{|c|}{\scriptsize{Dataset}} & \tiny{USPS1500} & \tiny{USPS1200} & \tiny{USPS900} & \tiny{USPS800} & \tiny{USPS700} \cr\hline
%  \multirow{10}{*}{\rotatebox{90}{\tiny{Number of samples in each category}}} & 0 & 1500 & 1200 & 900 & 800 & 700 \cr\cline{2-7}
%  & 1 & 1200 & 1200 & 900 & 800 & 700 \cr\cline{2-7}
%  & 2 & 900 & 900 & 900 & 800 & 700 \cr\cline{2-7}
%  & 3 & 800 & 800 & 800 & 800 & 700 \cr\cline{2-7}
%  & 4 & 800 & 800 & 800 & 800 & 700 \cr\cline{2-7}
%  & 5 & 700 & 700 & 700 & 700 & 700 \cr\cline{2-7}
%  & 6 & 800 & 800 & 800 & 800 & 700 \cr\cline{2-7}
%  & 7 & 700 & 700 & 700 & 700 & 700 \cr\cline{2-7}
%  & 8 & 700 & 700 & 700 & 700 & 700 \cr\cline{2-7}
%  & 9 & 800 & 800 & 800 & 800 & 700 \cr\hline
%  \multicolumn{2}{|c|}{UI} & 0.985 & 0.991 & 0.998 & 0.999 & 1 \\
%  \hline
%\end{tabular}
%\end{center}
%\end{table}
\begin{table*}
\caption{Performance of various clustering methods on USPS700. The best results are highlighted in \textbf{bold}. Our method comprehensively outperforms all the comparison methods on three metrics}
\label{tab:usps700}       % Give a unique label
\begin{center}
\begin{tabular}{|c|c|c|c|c|c|c|c|c|c|}
\hline
\multirow{2}{*}{Method} & \multicolumn{3}{c|}{ACC} & \multicolumn{3}{c|}{NMI} & \multicolumn{3}{c|}{ARI} \cr\cline{2-10}
& \textit{Min} & \textit{Max} & \textit{Med} & \textit{Min} & \textit{Max} & \textit{Med} & \textit{Min} & \textit{Max} & \textit{Med} \cr
\hline\hline
DDC~\cite{DBLP:journals/kbs/RenWLX20} & - & - & $0.967^*$ & - & - & $0.918^*$ & - & - & - \cr\hline
DDC-DA~\cite{DBLP:journals/kbs/RenWLX20} & - & - & $0.977^*$ & - & - & $0.939^*$ & - & - & - \cr\hline
DAC~\cite{Chang2017Deep} & 0.364 & 0.483 & 0.391 & 0.301 & 0.389 & 0.342 & 0.261 & 0.329 & 0.288 \cr\hline
K-means~\cite{6817617} & 0.469 & 0.564 & 0.534 & 0.367 & 0.451 & 0.433 & 0.259 & 0.368 & 0.344 \cr\hline
DEC~\cite{xie2016unsupervised} & 0.605 & 0.748 & 0.696 & 0.626 & 0.717 & 0.682 & 0.464 & 0.626 & 0.572 \cr\hline
SpectralNet~\cite{shaham2018spectralnet} & 0.827 & 0.877 & 0.835 & 0.860 & 0.898 & 0.865 & 0.788 & 0.876 & 0.799 \cr\hline
JULE~\cite{yang2016joint} & 0.856 & 0.954 & 0.877 & 0.862 & 0.893 & 0.888 & 0.802 & 0.908 & 0.840 \cr\hline
DCCM~\cite{DBLP:conf/iccv/WuLWQLLZ19} & 0.153 & 0.328 & 0.293 & 0.134 & 0.246 & 0.201 & 0.056 & 0.332 & 0.119 \cr\hline
Ours & \textbf{0.933} & \textbf{0.985} & \textbf{0.981} & \textbf{0.936} & \textbf{0.978} & \textbf{0.971} & \textbf{0.885} & \textbf{0.967} & \textbf{0.957} \cr
\hline
\end{tabular}
\end{center}
\end{table*}

%\begin{figure}
%  \centering
%  % Requires \usepackage{graphicx}
%  \includegraphics[width=7cm]{stability_usps.pdf}\\
%  \caption{Stability comparison among different methods. Evaluated on USPS700. The standard deviations of the performance (ACC, NMI, and ARI) of ten runs are illustrated.}\label{fig:stability_usps700}
%\end{figure}
\subsection{When intrinsicness is corrupted}
Cifar-10~\cite{krizhevsky2009learning} is a common benchmark for verifying object recognition algorithms. ImageNet-10~\cite{Chang2017Deep} is a tiny version of the original ImageNet dataset~\cite{deng2009imagenet}. In this subsection, we conduct experiments on these two datasets for examining if our method can deal with real-world object recognition tasks. We use the same architecture and hyperparameter configuration as Section \ref{sec:mnist} for experiments, except that we append a deconvolutional layer to the generator (correspondingly, a convolutional layer is appended to the discriminator and classifier respectively) to process 64x64 images (we resize the images in ImageNet-10 to 64x64). The experimental results are shown in Table \ref{tab:cifar10}. The results on Cifar-10 and ImageNet-10 are far from satisfactory. However, we can see that all the comparison methods (except for DCCM~\cite{DBLP:conf/iccv/WuLWQLLZ19}) have also failed on these two dataset. This illustrates that there is still a long way to go to apply unsupervised methods in object recognition tasks.

Three reasons may attribute to the poor performance of deep clustering in object recognition: First, object recognition itself is a challenging task. Understanding the class structure on these datasets requires a lot of out-of-domain knowledge. For examples, although the appearance of chickens, ostriches and canaries varies greatly, they are all considered as birds in the dataset. The same case happens on freighters, cruise ships and motorboats. In fact, if you ignore the background information, you will find that motorboats look more like cars, but motorboats are classified as ships along with freighters and cruise ships. Second, the quality of the generated samples are still poor, which undermines the criterion that the samples fetched from all clusters are equal to the original samples. This may be due to the insufficient capacity of the generator. Third, the images vary greatly in appearance. There may be many features that can be easily exploited as cues for cluster assignment, such as style and hue, not just the type of the objects. Therefore, when we introduce knowledge, such as invariance to translation, rotation, resize, brightness, contrast, saturation, hue, noise and etc., into the training process, we can further improve the clustering performance. That has been illustrated as Ours$^*$ in Table \ref{tab:cifar10}. Because this paper aims to build a general-purpose deep clustering framework, elaborating on exotic designs dedicated to extracting effective features for specific clustering tasks is beyond the scope of discussion. Here, we just make an indicative specification about incorporating any orthogonal techniques into the framework.

\begin{table}
\caption{Performance comparison on Cifar-10 and ImageNet-10. The best results are highlighted in \textbf{bold}. All the results of the comparison methods are reported in literature. For our method, the median statistics in 10 runs are reported}
\label{tab:cifar10}       % Give a unique label
\begin{center}
\begin{tabular}{|c|ccc|ccc|}
\hline
Datasets & \multicolumn{3}{c|}{Cifar-10} & \multicolumn{3}{c|}{ImageNet-10} \cr\hline
Method & ACC & NMI & ARI & ACC & NMI & ARI \cr\hline
K-means~\cite{6817617} & 0.229 & 0.087 & 0.049 & 0.241 & 0.119 & 0.057 \cr\hline
SC~\cite{zelnik2005self} & 0.247 & 0.103 & 0.085 & 0.274 & 0.151 & 0.076  \cr\hline
AC~\cite{gowda1978agglomerative} & 0.228 & 0.105 & 0.065 & 0.242 & 0.138 & 0.067  \cr\hline
NMF~\cite{cai2009locality} & 0.190 & 0.081 & 0.034 & 0.230 & 0.132 & 0.065  \cr\hline
AE~\cite{bengio2007greedy} & 0.314 & 0.239 & 0.169 & 0.317 & 0.210 & 0.152  \cr\hline
SAE~\cite{ng2011sparse} & 0.297 & 0.247 & 0.156 & 0.325 & 0.212 & 0.174  \cr\hline
DAE~\cite{vincent2010stacked} & 0.297 & 0.251 & 0.163 & 0.304 & 0.206 & 0.138  \cr\hline
DeCNN~\cite{5539957} & 0.282 & 0.240 & 0.174 & 0.313 & 0.186 & 0.142 \cr\hline
SWWAE~\cite{zhao1506stacked} & 0.284 & 0.233 & 0.164 & 0.324 & 0.176 & 0.160  \cr\hline
GAN~\cite{radford2015unsupervised} & 0.315 & 0.265 & 0.176 & 0.346 & 0.225 & 0.157  \cr\hline
JULE~\cite{yang2016joint} & 0.272 & 0.192 & 0.138 & 0.300 & 0.175 & 0.138  \cr\hline
DEC~\cite{xie2016unsupervised} & 0.301 & 0.257 & 0.161 & 0.381 & 0.282 & 0.203  \cr\hline
DAC~\cite{Chang2017Deep} & 0.522 & 0.396 & 0.306 & 0.527 & 0.394 & 0.302  \cr\hline
DCCM~\cite{DBLP:conf/iccv/WuLWQLLZ19} & \textbf{0.623} & \textbf{0.496} & \textbf{0.408} & \textbf{0.710} & \textbf{0.608} & \textbf{0.555}  \cr\hline
Ours & 0.330 & 0.315 & 0.014 & 0.368 & 0.377 & 0.030  \cr
Ours$^*$ & 0.440 & 0.421 & 0.223 & 0.487 & 0.492 & 0.310  \cr
\hline
\end{tabular}
\end{center}
\end{table}

%\begin{figure}
%  \centering
%  % Requires \usepackage{graphicx}
%  \includegraphics[width=8cm]{cifar10.pdf}\\
%  \caption{Learning curves on cifar-10. The same learning curves are depicted as that in Fig.\ref{fig:mnist} and Fig.\ref{fig:fashionmnist}. The generated samples are also arranged according to the class labels. The samples of each row in the figure are generated from the same class label. The optimization is performed by 100000 iterations in total.}\label{fig:cifar10}
%\end{figure}
\subsection{Complexity analysis}
The training of the framework consumes a relatively long time (generally, 38 hours for MNIST, Fashion-MNIST, Artifact-MNIST and USPS, 1.2 hours for ORL, 56 hours for Cifar-10, 220 hours for ImageNet-10), but after training, the framework outputs the clustering result for an instance (32x32 grayscale/color image or 64x64 color image) within 0.03/0.034/0.09 milliseconds (average over a batch size of 50)\footnote{All the experiments are performed on computer with Ubuntu 18.04.1 LTS, Intel(R) Xeon(R) CPU E5-1620 v4 @ 3.50GHz, NVIDIA GeForce GTX 1080 Ti, CUDA 10.0, Python 3.6, Tensorflow 1.2.}. It should be noted that since we used similar experimental configurations for all the datasets, the reported performance must be below its maximum value. In this case, when a better hyperparameter is selected, the measured training latency can be reduced by cutting the number of iterations.

\section{Conclusions and Future Work}\label{sec:Con}
In this paper, we have defined the objective of deep clustering as finding a precise feature as the cue for cluster assignment. To achieve this objective, we proposed a general-purpose deep clustering framework that integrates representation learning and clustering into a single pipeline for joint optimization. We applied the proposed framework to a synthetic dataset and several real-world image benchmarks. The results showed that the framework performed better than, or comparably to, the baselines. We attribute the promising results to the fact that our framework captures the intrinsic characteristics of samples and learns to select one whose discrete value space exactly matches the specified categories as the cue for cluster assignment.

However, there are still some limitations for the proposed framework. First, the uniform prior imposed on the clusters only benefits when the samples are approximately uniformly distributed across clusters. Second, the failure on the object recognition datasets suggests that pure statistical methods are difficult to solve complex recognition problems, and it is necessary to introduce additional knowledge or visual mechanisms into the unsupervised framework. In the future, we aim to build more robust methods in order to cater for various clustering scenarios, including adopting a learnable prior to fit more general distributions and using the cognition knowledge provided by humans to induce the learning procedure.

\section*{Acknowledgements}
The code and data are available on \url{https://github.com/gyh5421/unified_deep_clustering}.
\appendices

\ifCLASSOPTIONcaptionsoff
  \newpage
\fi



\bibliographystyle{IEEEtran}
\bibliography{IEEEabrv,IEEEexample}
%\begin{thebibliography}{1}
%\end{thebibliography}

\vspace{-12 mm}
\begin{IEEEbiography}[{\includegraphics[width=1in,height=1.25in,clip,keepaspectratio]{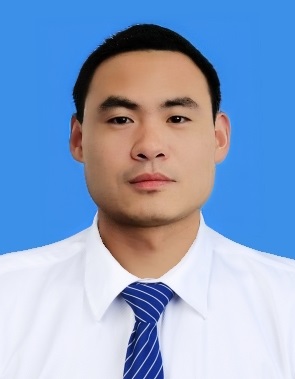}}]{Yanhai Gan}
received the B.Eng. and M.Eng. degrees from the Department of Computer Science and Technology, Ocean University of China, Qingdao, China, in 2014 and 2017, respectively.

He is currently a Ph.D. student in the School of Computer Science and Technology, Ocean University of China. His research interests include machine learning, pattern recognition and image processing.
\end{IEEEbiography}
\vspace{-14 mm}
\begin{IEEEbiography}[{\includegraphics[width=1in,height=1.25in,clip,keepaspectratio]{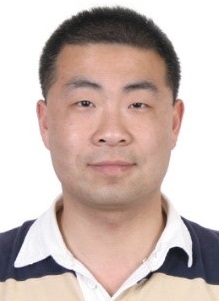}}]{Xinghui Dong}
received the Ph.D. degree from Heriot-Watt University, U.K., in 2014. He has worked as a research associate with the Centre for Imaging Sciences, the University of Manchester, U.K.. Then he joined the Ocean University of China in 2021. His research interests include automatic defect detection, image representation, texture analysis, and visual perception.
\end{IEEEbiography}
\vspace{-14 mm}
\begin{IEEEbiography}[{\includegraphics[width=1in,height=1.25in,clip,keepaspectratio]{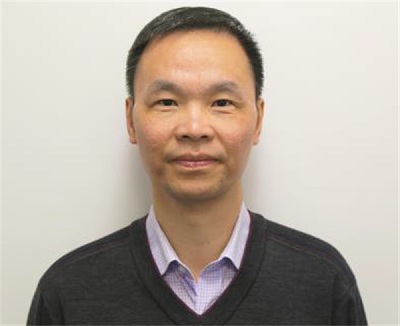}}]{Huiyu Zhou}
received the B.Eng. degree in Radio Technology from Huazhong University of Science and Technology of China and the M.Sc. degree in Biomedical Engineering from University of Dundee of United Kingdom. He was awarded the Ph.D. degree in Computer Vision from Heriot-Watt University, Edinburgh, United Kingdom. He is a Professor and currently heads the Applied Algorithms and AI (AAAI) Theme at University of Leicester. Prof. Zhou has published widely in the field.
\end{IEEEbiography}
\vspace{-14 mm}
\begin{IEEEbiography}[{\includegraphics[width=1in,height=1.25in,clip,keepaspectratio]{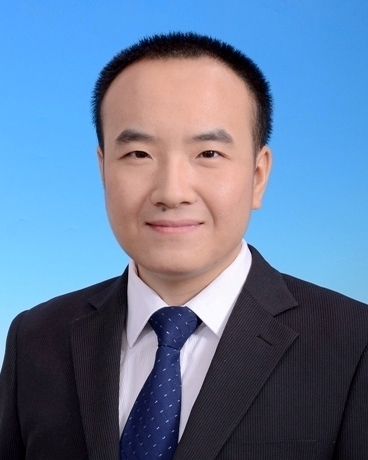}}]{Feng Gao}
received the B.Sc. degree in software engineering from Chongqing University, Chongqing, China, in 2008, and the Ph.D. degree in computer science and technology from Beihang University, Beijing, China, in 2015.

He is currently an Associate Professor with the School of Computer Science and Technology, Ocean University of China. His research interests include remote sensing image analysis, pattern recognition and machine learning.
\end{IEEEbiography}
\vspace{-14 mm}
\begin{IEEEbiography}[{\includegraphics[width=1in,height=1.25in,clip,keepaspectratio]{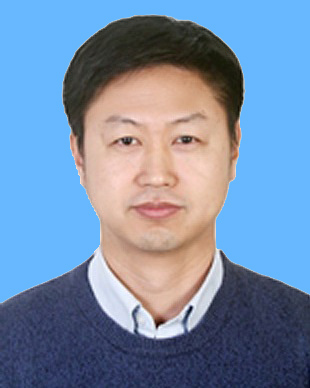}}]{Junyu Dong}
received the B.Sc. and M.Sc. degrees from the Department of Applied Mathematics, Ocean University of China, Qingdao, China, in 1993 and 1999, respectively, and the Ph.D. degree in image processing from the Department of Computer Science, Heriot-Watt University, Edinburgh, U.K., in 2003.

He is currently a Professor, the Dean of School of Computer Science and Technology and the Dean of Haide College, Ocean University of China. His research interests include visual information analysis and understanding, machine learning and underwater image processing.
\end{IEEEbiography}

\end{document}